\PassOptionsToPackage{table}{xcolor}
\documentclass[sigconf]{acmart}
\copyrightyear{2025}
\acmYear{2025}
\setcopyright{cc}
\setcctype{by}
\acmConference[CHI '25]{CHI Conference on Human Factors in Computing Systems}{April 26-May 1, 2025}{Yokohama, Japan}
\acmBooktitle{CHI Conference on Human Factors in Computing Systems (CHI '25), April 26-May 1, 2025, Yokohama, Japan}
\acmDOI{10.1145/3706598.3713329}
\acmISBN{979-8-4007-1394-1/25/04}

\usepackage{geometry}
\usepackage{cleveref}
\usepackage{caption}
\usepackage{subcaption}
\usepackage{tabularx}
\usepackage{makecell}
\usepackage{cellspace}
\usepackage{ragged2e}
\usepackage{booktabs}
\usepackage{multirow}

\graphicspath{ {./figures/} }

\newcommand{\heading}[1]{\multicolumn{1}{c}{#1}}
\definecolor{Gray}{gray}{0.9}
\newcolumntype{T}{>{\RaggedRight\arraybackslash}X}
\newcolumntype{Y}{>{\Centering\arraybackslash}X}
\newcolumntype{G}{>{\Centering\arraybackslash\columncolor{Gray}}X}
\begin{document}

\title[]{\texorpdfstring{Perceptions of Sentient AI and Other Digital Minds: \\ Evidence from the AI, Morality, and Sentience (AIMS) Survey}{Perceptions of Sentient AI and Other Digital Minds: Evidence from the AI, Morality, and Sentience (AIMS) Survey}}

\author{Jacy Reese Anthis}
\orcid{0000-0002-4684-348X}
% \affiliation{
%   \institution{Sentience Institute}
%   \city{New York}
%   \country{USA}
% }
% \affiliation{
%   \institution{Stanford University}
%   \city{Stanford}
%   \country{USA}
% }
\affiliation{
  \institution{University of Chicago}
  \city{Chicago}
  \country{USA}
}
\email{jacy@sentienceinstitute.org}

\author{Janet V.T. Pauketat}
\orcid{0000-0003-3280-3345}
\affiliation{
  \institution{Sentience Institute}
  \city{New York}
  \country{USA}
}
\email{janet@sentienceinstitute.org}

\author{Ali Ladak}
\orcid{0000-0003-1039-5774}
% \affiliation{
%   \institution{Sentience Institute}
%   \city{New York}
%   \country{USA}
% }
\affiliation{
  \institution{University of Edinburgh}
  \city{Edinburgh}
  \country{UK}
}
\email{ali@sentienceinstitute.org}

\author{Aikaterina Manoli}
\orcid{0000-0003-2562-0380}
% \affiliation{
%   \institution{Sentience Institute}
%   \city{New York}
%   \country{USA}
% }
\affiliation{
  \institution{Max Planck Institute for Human Cognitive and Brain Sciences}
  \city{Leipzig}
  \country{Germany}
}
\email{katerina@sentienceinstitute.org}

\begin{abstract}
  Humans now interact with a variety of \textit{digital minds}, AI systems that appear to have mental faculties such as reasoning, emotion, and agency, and public figures are discussing the possibility of sentient AI. We present initial results from 2021 and 2023 for the nationally representative AI, Morality, and Sentience (AIMS) survey ($N$ = 3,500). Mind perception and moral concern for AI welfare were surprisingly high and significantly increased: in 2023, one in five U.S. adults believed some AI systems are currently sentient, and 38\% supported legal rights for sentient AI. People became more opposed to building digital minds: in 2023, 63\% supported banning smarter-than-human AI, and 69\% supported banning sentient AI. The median 2023 forecast was that sentient AI would arrive in just five years. The development of safe and beneficial AI requires not just technical study but understanding the complex ways in which humans perceive and coexist with digital minds.
\end{abstract}



\hyphenpenalty=50
\hyphenpenalty=1000

\maketitle

\section{Introduction}

Philosophers and scientists have long considered the possibility of artificial intelligence (AI) systems with mental faculties such as reasoning, emotion, and agency, which we call \textit{digital minds}. Popular questions include: Can an AI ever have a mind of its own? How should we treat sentient AI if it is created? Should humanity, as \citet{metzinger21} suggests, ban the development of sentient AI to preclude interaction? When, if it is not banned, should we expect sentient AI to first be created? These profound questions of mind perception, moral status, policy, and forecasting are now being asked and answered by researchers, industry leaders, and much of the general public.\footnote{Up-to-date results for the AI, Morality, and Sentience (AIMS) survey are available at \url{https://sentienceinstitute.org/aims-survey}.}

\begin{figure}[!t]
    \includegraphics[width=\linewidth]{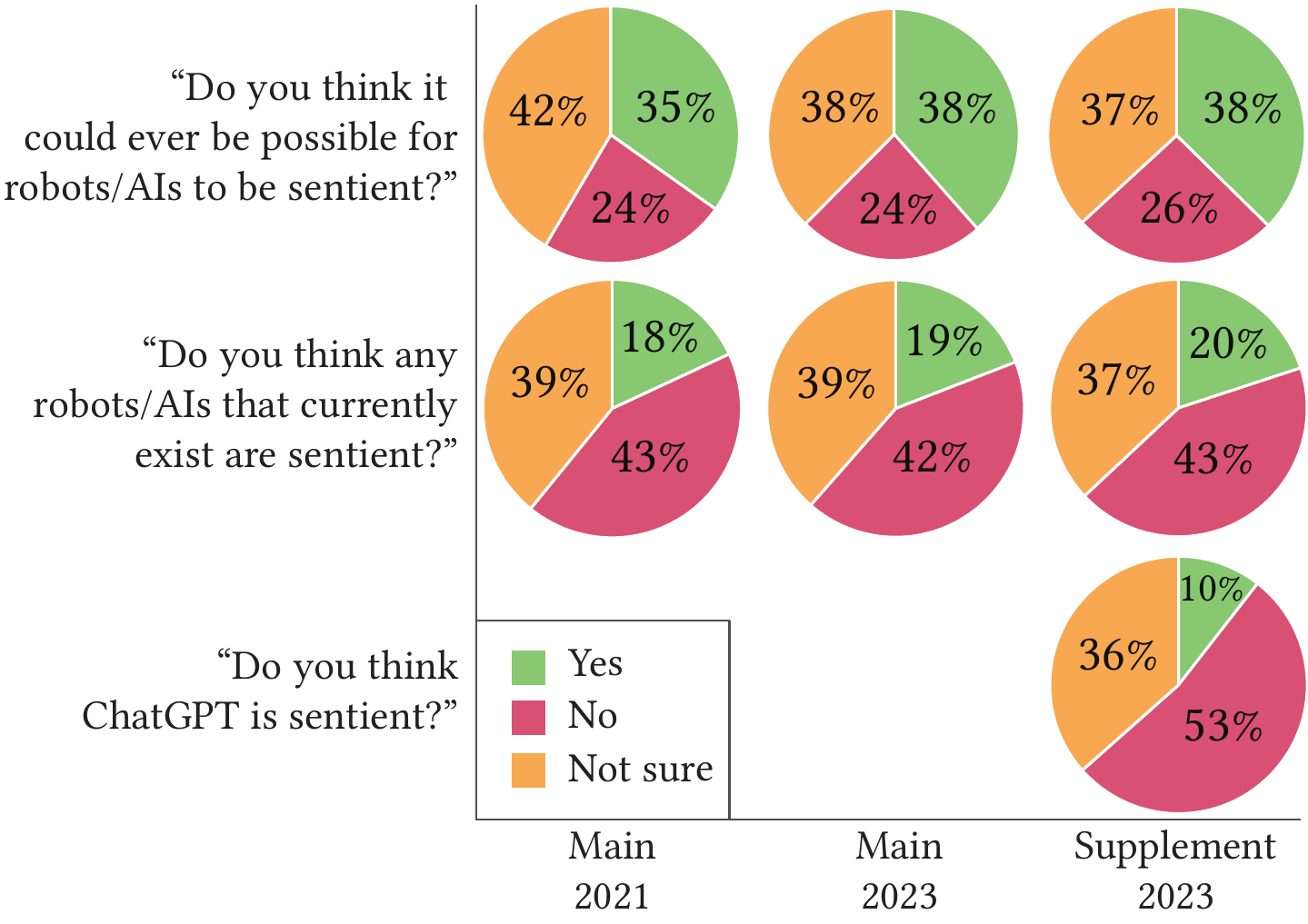}
    \caption{Responses to three questions across the three AIMS survey waves. See \Cref{sec:results} for detailed results.}
    \label{fig:sentience}
    \Description{A 3-by-3 grid with three questions as rows: “Do you think it could ever be possible for robots/AIs to be sentient?”, “Do you think any robots/AIs that currently exist are sentient?”, and “Do you think ChatGPT is sentient?”; and three survey waves as columns: Main 2021, Main 2023, and Supplement 2023. Only the Supplement 2023 asked the third question.}
\end{figure}

The manifestation of these dynamics in public discourse surged in 2022, first in February when Ilya Sutskever, chief scientist at a leading AI company, OpenAI, tweeted, “it may be that today’s large neural networks are slightly conscious” \cite{sutskever22}. In June, Google vice-president Blaise Agüera y Arcas published an editorial in \textit{The Economist} on how “Artificial neural networks are making strides towards consciousness” \cite{aguerayarcas22}. Weeks later, Google engineer Blake Lemoine vigorously argued that the company’s chatbot LaMDA was sentient and needed legal protection, which led to his termination from the company and sparked a global conversation on the topic \cite{tiku22}. Since before 2022, there has also been rapid growth in the academic literature on the possibility of digital minds and the normative implications \cite{harris21}.

Regardless of whether sentient AI ever will be developed, the act of asking and answering questions about digital minds will itself reshape human-AI interaction. In this work, we view sentience and consciousness not as facts of reality but as semantic labels that humans place on certain entities, endowing them social status \cite{anthis22}. We know from the human-computer interaction (HCI) and human-robot interaction (HRI) literatures that “computers are social actors” (CASA) \cite{hou23, nass94}; are perceived as having their own minds \cite{gray07, scott23, stuart21, thellman22, wang21b}; and can be perceived as moral patients or subjects with their own welfare \cite{freier08, harris21, kahn04, lima20, pauketat22} and as moral agents worthy of praise, blame, and responsibility \cite{banks19a, freier07, jackson19, kneer21a}. For example, \citet{scott23} conducted an online survey of 100 Amazon Mechanical Turk workers on perceptions of machine consciousness and identified dynamic tensions in user perceptions, and \citet{kneer21a} found that whether people saw a harmful robot as blameworthy depended on whether it seemed to have its own theory of mind.

Perceptions of AI already shape trust in AI \cite{hoff15}, willingness to use AI \cite{kelly23, kim22a, tiwari23}, and the consequences of AI for mental health and social relationships \cite{cho24, hohenstein20, hohenstein23, park24}. In the long run, human perceptions of and interactions with digital minds will affect which types of AI systems are safe, which policy and governance frameworks are most beneficial, how AI designers and engineers choose to build AI, how individual users and state actors use AI, and ultimately the existential trajectory of the human species. A particularly concerning and understudied existential risk \cite{bostrom14, good65, russell19} is the effect on human agency like the dystopian futures in the 1909 short story \textit{The Machine Stops} and the science fiction film \textit{WALL-E}. If humans fail to prepare for these new forms of human-AI interaction, we could face “disempowerment,” absent-mindedly giving up control to evolutionary or artificial forces \cite{dung24, eisenpress24, fernandez24, grace24, kulveit25, salib24}, and this transition may be particularly difficult to preempt if it is “accumulative” \cite{kasirzadeh25} or “gradual” \cite{kulveit25}.

Effective AI research and design require grounded knowledge of the different “frames,” schema of interpretation \cite{goffman74}, that increasingly shape how people individually and collectively understand and react to AI systems. Representative surveys to tease out such understandings are a well-established method in HCI, such as in privacy \cite{haring23}, cybercrime \cite{gupta24}, and trust in social media platforms \cite{zhang24a}. Attitudes and beliefs have implications for technological design, including privacy \cite{chapman22, haring23, herbert23}, cybercrime \cite{breen22, gupta24}, and trust in social media \cite{zhang24}. Theories in psychology and sociology emphasize the distinct role of sociocognitive processes in shaping the future, such as through “world-making” \cite{pauketat25, power23}, and a wide range of research has shown that public opinion has influence on policy, politics, and other social institutions \cite[e.g.,][]{barbera19, burstein03}.

We address mind perception, morality, policy, and forecasting by posing four overlapping research questions (RQs) about public perceptions of digital minds:

\begin{itemize}
    \setlength\itemsep{0.5em}
    \item \textbf{RQ1}: To what extent do people perceive sentience and other mental faculties in AI?
    \item \textbf{RQ2}: To what extent do people feel morally concerned for AI (i.e., see it as a moral subject) and threatened by AI (i.e., see it as a moral agent)?
    \item \textbf{RQ3}: What policies are supported to govern interaction between humans and sentient AI?
    \item \textbf{RQ4}: When do people expect sentient AI to arrive, and what do they think will happen when it does?
\end{itemize}

Because there have been such rapid changes in how AI is discussed, used, and conceptualized by the public in the 21\textsuperscript{st} century, we began seeking answers to these questions in 2021 with the first wave of the \textbf{AI, Morality, and Sentience (AIMS) survey} with a nationally representative sample of U.S. adults. In this paper, we present the first three waves of AIMS: one in 2021 and two in 2023 (one with the same questions as the 2021 wave and one with supplemental questions).

Our first wave (\textit{N} = 1,232, referred to as \textbf{Main 2021}), was collected from November to December 2021, approximately seven months before the topic of sentient AI entered the public spotlight with the LaMDA discussion. We asked 86 questions related to the perception of mental faculties in AI, the moral concern for AI being harmed, the moral threat of AI enacting harm, the prospect of banning sentience-related technologies, forecasts of the future of sentient AI, personal AI usage, demographics, and relevant background beliefs such as views on animal welfare and environmental issues. From April to May 2023, approximately 11 months after the public spotlight began and four months after the public release of ChatGPT—an AI product that became the world’s fastest growing app \cite{hu23}—we repeated the AIMS survey with a new nationally representative sample (\textit{N} = 1,169, referred to as \textbf{Main 2023}). Finally, to assess a wider range of attitudes and better contextualize the longitudinal results, we conducted a supplemental AIMS survey wave (\textit{N} = 1,099, referred to as \textbf{Supplement 2023}) from May to July 2023 with 111 questions, including on public awareness of specific AI systems, trust of AI and its developers, and additional forecasts of the future of artificial general intelligence (AGI). We intend AIMS supplements to be one-time surveys that do not necessarily measure the same perceptions over time but complement the longitudinal main data.

In \textbf{Main 2021}, we found surprisingly high mind perception (\textbf{RQ1}) and attribution of moral status, including both concern for the wellbeing of AI and feeling threatened by AI (\textbf{RQ2}). Likewise, even the most strongly worded policy proposals garnered substantial support, such as 37.2\% agreement with, “I support granting legal rights to sentient robots/AIs” (\textbf{RQ3}). Participants forecasted sentient AI would arrive quickly with a median estimate of five years (\textbf{RQ4}). Comparing data from \textbf{Main 2021} to \textbf{Main 2023}, there were significant increases in mind perception, moral concern, and threat as well as a shortening timeline of when sentient AI will arrive. In a set of additional policy questions asked in \textbf{Supplement 2023}—prompted by the political debates in early 2023 on the topic—we also found widespread support for slowing down and regulating various forms of AI development. Finally, as described in the supplemental materials, exploratory analysis surfaced numerous significant differences in opinion across age, gender, frequency of AI interaction, and other participant characteristics, even when adjusting for family-wise error rates.

These preliminary results from the ongoing AIMS survey project suggest a number of implications for HCI design, policy, and research. In \Cref{sec:discussion}, we argue that:

\begin{enumerate}
    \item Designers should prioritize explainable AI (XAI) and consider selectively tuning anthropomorphism with physical and behavioral cues to harness benefits and minimize risks of different systems.
    \item Policymakers should prioritize safety-focused policy, such as the EU AI Act, to account for widespread public concern. They should also ensure their own technological literacy and work towards meaningful public engagement with these challenging new issues.
    \item Researchers should address the open and important research questions about the perception of sentient AI, including the drivers and consequences of public opinion, refinement of HCI theories, and approaches to varied global perspectives on AI development.
\end{enumerate}

Understanding the “theory of mind” that people use to think about increasingly powerful and prevalent AI systems can illuminate the design space of future human-AI interaction and inform the management of existing AI technologies. Rigorously measuring public perceptions, particularly over time, can provide insights for design and policy decisions that must be made before new technologies are invented and that should account for what people value, not only how they behave with existing technologies. Survey data is particularly important during this challenging and dynamic period of technological change as the HCI community debates the interplay between new forms of human-AI interaction and established areas of research, such as CSCW and social computing \cite{morris24}.

\section{Related Work}

\subsection{Social Response and Mind Perception}

There are numerous mechanisms by which humans interact with computer systems in ways similar to social interactions between humans. These span the course of interaction from initial perception of the system to attitude and belief formation to resultant behavior. The CASA view, as evidenced by studies in the 1990s, suggests that these “social” responses need not occur because of a conscious belief that the computer has human characteristics but can merely be the application of etiquette, stereotypes, norms, and other social scripts \cite{nass94, reeves96}. Recent work has continued to build on social response theory to explain human-AI interaction, such as with chatbots \cite{nguyen23} and avatars \cite{miao22}. As computers have become more familiar and ubiquitous, interface design has built on this tendency by incorporating natural social dynamics between the user and the system \cite{shneiderman17}, and now people also use novel “human-media social scripts” \cite{gambino20}. For example, with voice assistants, such as Amazon Alexa, people often have established social routines and commands that have been developed through repeated usage and understanding of the system’s affordances \cite{ammari19}. “Mindless” social responses can emerge from habit and ease of use \cite{nass00}, and studies continue to find that many social psychology effects found in human-human interaction carry over to HCI \cite{ladak23c, srinivasan16}.

As computer systems have become more sophisticated, they have in many ways become more human-like, leading to a research focus on anthropomorphism (i.e., the attribution of human characteristics, motivations, intentions, or emotions to nonhuman entities \cite{epley07}). While anthropomorphism is often associated with positive HCI outcomes through increased trust and engagement, it can lead to unrealistic expectations \cite{luger16} and the well-known “uncanny valley” effect \cite{strait15}.

A particularly important human-like feature that people can attribute is having a mind. The importance of mind perception in HCI has led to a diverse, interdisciplinary literature that uses a number of different, closely related terms, including “theory of mind” and “mind attribution,” as documented in a recent HRI review \cite{thellman22}. Several authors have taxonomized the mental faculties perceived in human and nonhuman entities: \citet{gray07} used principal component factor analysis to identify “experience” (e.g., hunger, fear, pain, pleasure) and “agency” (e.g., self-control, morality, memory, emotion recognition) as the main dimensions of mind perception. More recent work has proposed one-dimensional \cite{tzelios22}, three-dimensional \cite{weisman17}, and five-dimensional taxonomies \cite{malle19a}. In the HCI literature, \citet{wang21b} recently proposed “mutual theory of mind” as a framework to study human-AI interaction and to design systems that support user needs over time. Related literature also suggests that techno-animism—the attribution of souls, life forces, and related phenomena to technological artifacts—plays an important role in human-AI interaction \cite{pauketat22, richardson16, seymour20}. Anthropomorphic cues, in particular those affecting mind perception, have been shown to shape beliefs about system performance and moral responsibility in contexts such as driving simulators \cite{waytz14}.

\subsection{Sentience and Morality}

Social response and mind perception are key drivers of moral attitudes and behaviors. \citet{gray12} aphorized, “Mind perception is the essence of morality.” Sentience has been one of the most frequently hypothesized and debated mental faculties of future AI in science fiction and purportedly even current AIs, as attested in the statements of OpenAI's \citet{sutskever22} and Google's \citet{aguerayarcas22} and Lemoine \cite{tiku22}. Sentience (i.e., the capacity for positive and negative experiences \cite{anthis21}) is closely associated with moral standing as it is frequently seen as the primary or exclusive basis of moral concern \cite{perry24}. In the two-dimensional taxonomy of mind as experience and agency, perceptions of sentience are typically associated with experience but can also affect perceived agency, particularly the capacity to do harm (i.e., to pose a threat). Sometimes people refer to “sentience” with the term “consciousness,” but while consciousness has many different meanings in the literature, we focus on sentience for specificity and to focus on social and moral entailments \cite{ladak23c}. We also use the term “digital mind” for the wider category—to encompass AI not necessarily with sentience but any mental faculties, such as reasoning \cite{huang23} or understanding \cite{bender20}.

In terms of moral concern for an AI's wellbeing (i.e., the attribution of moral standing), \citet{harris21} conducted a systematic literature review of empirical studies of moral concern for AIs, finding that AI tends to be granted much less moral concern than humans, but that moral concern increases when humans perceive autonomy, human-like appearance, mind, and verbal responses to harm in the AI. For example, \citet{spence18} presented 167 undergraduate students with a video of either a human or a robot asking them to sign a petition for robot rights. Responses did not depend on whether the requester was a human or robot, and 46\% said they would sign the petition—a substantial amount of moral concern—and perceived credibility, positive attitudes towards robots, and prior experience with robots was associated with a higher likelihood of signing the petition. Similarly, \citet{lima20} presented 11 possible AI rights and corresponding arguments to debunk common misconceptions about them to 1,270 Amazon Mechanical Turk users and 164 Qualtrics users. Participants tended to disagree with the endowment of most rights, such as the right to enter contracts and the freedom of speech, with the exception of protection from cruel treatment. Each participant was randomly assigned to one of four debunking interventions, which were found to significantly increase support for AI rights, particularly the intervention that showed examples of nonhumans that were already granted rights and duties, which indicated to participants that AI rights were a realistic possibility.

In terms of being threatened by AI (i.e., the attribution of negative moral agency), this has been most salient with AI that assists in consequential tasks, such as medical imaging \cite{lebovitz21} and bail decisions \cite{grgic-hlaca19, lima21}. AI is often blamed or held responsible when they cause harm \cite{freier09, friedman95, kneer21a, komatsu16, komatsu21, stuart21, tolmeijer22}, though there have been mixed results in studies of whether they are blamed more or less than humans in similar contexts \cite{lee21a, lima23, malle19}. While many scholars have developed theories of artificial moral agency in ethics and HCI \cite{cervantes20, zoshak21}, others have argued that machines themselves lack minds or personhood and therefore should not—or cannot—be held responsible for harm and that this attribution can be a distraction from holding the humans involved accountable for their actions \cite{sparrow07}. A rapidly growing academic and public discourse addresses the existential threat that “agentic,” “conscious,” or “superintelligent” AI poses to humanity itself, such as if capabilities rapidly accelerate in an “intelligence explosion” \cite{bostrom14, good65, russell19}. Such threats have been considered to some extent for decades, particularly in science fiction, but they captured the public imagination anew following the aforementioned mid-2022 discussions of sentient AI, the launch of ChatGPT in November 2022, and the widely read open letter signed by leading academics and executives in March 2023 that called for a 6-month pause on advanced AI development \cite{futureoflifeinstitute23}.

The field of HCI lacks a solid foundation of detailed, longitudinal, and nationally representative data on which social theories and technical designs can be developed. Basic facts about everyday user viewpoints, such as whether people support protecting sentient AI from harm or banning the development of sentient AIs, remain unknown. Therefore, in the present work, we answer our RQs with rigorous, multidimensional measurement of the beliefs and attitudes toward morality and sentience that could drive human-AI interaction in the 21\textsuperscript{st} century.

\subsection{Perceptions and Public Opinion}

Internal states (perceptions, opinions, beliefs, attitudes, etc.) drive individual behavior and social change. These cognitive and affective states deserve study in their own right, not just as proxies for behavior. There are large and well-documented differences between them and behavior, often summarized as the “attitude-behavior gap” \cite{boulstridge00}. For example, it is well-known that people report significant concerns about preserving their own privacy, yet they tend to take little action to preserve it with their online behavior; this “privacy paradox” has motivated efforts to preserve privacy without requiring individual behavioral choices—in an effort to align outcomes with attitudes rather than merely with behaviors \cite{solove20, yang24}.

Instead of as a proxy for behavior, cognition is studied for its distinct causal effects on future real-world events. That can be through interaction with other cognitive processes of the individual mind or as a distinct driver of subsequent behavior, such as speaking with peers, that affects other people or the environment. For example, the expression of political views and the publication of poll results can affect how people vote, among other real-world outcomes \cite{roy15, roy21}. Normatively, it is often the case that, if a person could decide whether their attitude or behavior is used as the basis of a design or policy decision, they would choose the attitude, such as high proportions of people voting for recycling mandates despite few people opting in before it is mandated or the many people attempting to diet who pay a coach or use an app to push themselves to consistently change their behavior.

In the AIMS survey project, we draw on sociocognitive theories of how people think about different “frames,” the schema of interpretation by which people make sense of the world \cite{goffman74}. In particular, people collectively engage in “world-making,” discussed in more detail in the context of sentient AI in \citet{pauketat25}. World-making is the process by which individuals, including researchers, envision possible futures such as utopias and dystopias and bring one into existence \cite{pauketat25, power23, savage24}. In general, a central finding of social psychology is that “people are future-oriented and often are guided more by what could be than what is” \cite{power23}, and human motivation centers the pursuit of positive outcomes and, moreso, the avoidance of negative outcomes \cite{kahneman79}. Social discourse and debate often center “collective imagination” \cite{borer10} or different “imaginaries” \cite{augustine19, sartre72} of how the world could be. Likewise, design fiction \cite{dunne13} and speculative design \cite{dunne13, wyche22} encourage people to explore a variety of likely or unlikely possible futures.

\section{Methodology}
\label{sec:methodology}

In order to study change in public opinion over time, we have been collecting longitudinal data in the AI, Morality, and Sentience (AIMS) survey. In this paper, we report the first three waves of survey data: the main survey in 2021 (i.e., \textbf{Main 2021}), the main survey in 2023 with the same questions as in 2021 (i.e., \textbf{Main 2023}), and a supplemental survey in 2023 with a different set of questions (i.e., \textbf{Supplement 2023}). To allow for direct comparison between results, AIMS participants who had taken one survey were excluded from the following survey waves, and the samples were otherwise gathered with an identical methodology to ensure representativeness and that we could make statistical comparisons over time, which past one-time surveys on AI cannot provide. We intend to continue running the main AIMS survey and collecting different supplemental data over time as humanity begins to coexist with digital minds in the coming years.

\begin{table*}[htbp]
    \centering
    \caption{Unweighted demographics for the three AIMS survey waves: \textbf{Main 2021}, \textbf{Main 2023}, and \textbf{Supplement 2023}. Unweighted proportions are those before raking, which further balanced the data. Demographic categories (e.g., binary gender) were determined by the sample providers, as were population estimates, which were from the latest iteration available of the U.S. Census Bureau’s American Community Survey at the time of data collection.}
    \begin{tabularx}{\textwidth}{lYGYYG}
        \hline
        & \textbf{\multirowcell{3}[0pt]{2021 \\ Main}} & \textbf{\rule{0pt}{3ex} Latest U.S. \newline Population \newline Estimates} & \textbf{\multirowcell{3}[0pt]{2023 \\ Main}} & \textbf{\multirowcell{3}[0pt]{2023 \\ Supp.}} & \textbf{Latest U.S. \newline Population \newline Estimates} \\[6.5ex]
        \hline
        \textbf{Age} & & & & & \\
            \quad 18–34 & 22\% & 24\% & 27\% & 28\% & 28\% \\
            \quad 35–54 & 32\% & 31\% & 34\% & 33\% & 33\% \\
            \quad 55– & 45\% & 45\% & 39\% & 39\% & 39\% \\
            \textbf{Gender} & & & & & \\
            \quad Female & 54\% & 52\% & 54\% & 51\% & 51\% \\
            \quad Male & 46\% & 48\% & 46\% & 49\% & 49\% \\
            \textbf{Region} & & & & & \\
            \quad Midwest & 20\% & 21\% & 21\% & 21\% & 21\% \\
            \quad Northeast & 19\% & 18\% & 18\% & 17\% & 17\% \\
            \quad South & 40\% & 38\% & 39\% & 38\% & 38\% \\
            \quad West & 21\% & 24\% & 23\% & 24\% & 24\% \\
            \textbf{Household Income} & & & & & \\
            \quad –\$24,999 & 12\% & 12\% & 14\% & 12\% & 11\% \\
            \quad \$25,000–\$49,999 & 18\% & 18\% & 18\% & 16\% & 17\% \\
            \quad \$50,000–\$74,999 & 19\% & 19\% & 15\% & 17\% & 17\% \\
            \quad \$75,000–\$99,999 & 16\% & 16\% & 16\% & 14\% & 14\% \\
            \quad \$100,000– & 36\% & 36\% & 39\% & 40\% & 42\% \\
            \textbf{Ethnicity/Race} & & & & & \\
            \quad Asian & \phantom{0}6\% & \phantom{0}6\% & \phantom{0}6\% & \phantom{0}6\% & \phantom{0}6\% \\
            \quad Black & \phantom{0}12\% & 12\% & 11\% & 11\% & 11\% \\
            \quad Hispanic (any race) & 16\% & 17\% & 17\% & 17\% & 17\% \\
            \quad Native American & \phantom{0}1\% & \phantom{0}1\% & \phantom{0}0\% & \phantom{0}0\% & \phantom{0}0\% \\
            \quad White & 64\% & 64\% & 61\% & 61\% & 61\% \\
            \quad Other & \phantom{0}2\% & \phantom{0}2\% & \phantom{0}4\% & \phantom{0}4\% & \phantom{0}4\% \\
            \textbf{Education} & & & & & \\
            \quad Less than high school & \phantom{0}9\% & 10\% & 10\% & 10\% & 10\% \\
            \quad High school & 25\% & 27\% & 25\% & 26\% & 27\% \\
            \quad Some college & 20\% & 21\% & 17\% & 18\% & 21\% \\
            \quad Associate & 10\% & \phantom{0}9\% & 10\% & \phantom{0}9\% & \phantom{0}9\% \\
            \quad Bachelor's degree & 20\% & 20\% & 25\% & 22\% & 21\% \\
            \quad Post-graduate & 16\% & 13\% & 13\% & 13\% & 13\% \\
        \bottomrule
    \end{tabularx}
    \label{tab:demos}
    \Description{Unweighted demographics for the three survey waves: the main 2021 survey, the main 2023 survey, and the supplemental 2023 survey. Unweighted proportions are those before raking, which further balanced the data. Demographic categories (e.g., binary gender) were determined by the sample providers, as were population estimates, which were from the latest iteration available of the U.S. Census Bureau’s American Community Survey at the time of data collection. The rows of the table are demographic categories (e.g., Region: Midwest), and the columns are: 2021 Main, Latest U.S. Population Estimates as of 2021, 2023 Main, 2023 Supp., and Latest U.S. Population Estimates as of 2023. The data is available in CSV format in the supplemental materials.}
\end{table*}

\subsection{Recruitment and Census-Balanced Demographics}

Each of the three AIMS survey waves was conducted with a nationally representative sample of U.S. adults aged 18 or older. Participants were recruited through a combination of Ipsos iSay, Dynata, Disqo, and other leading survey panels to ensure representativeness. Sample sizes were initially targeted at 1,100 participants, corresponding to a ±3\% margin of error, and additional participants were recruited as needed to ensure representativeness of each subgroup. Unweighted sample proportions and U.S. adult population estimates are shown in \Cref{tab:demos}. Each sample was collected based on U.S. census data for age, gender, race/ethnicity, income, and education. To further ensure external validity, we report sample statistics (e.g., median, mean, standard error) that are weighted with iterative proportional fitting, a procedure commonly known as “raking” that adjusts sample weights to mitigate the random demographic variation that is present even in representative sampling \cite{deming40}.

To contextualize our results, we documented several AI-specific characteristics of the AIMS participants. In \textbf{Main 2023}, 29.2\% answered “Yes” to the binary question, “Do you own AI or robotic devices that can detect their environment and respond appropriately?” alongside examples, and 16.5\% answered “Yes” to “Do you work with AI or robotic devices at your job?” alongside another set of examples. We also measured smart device ownership, the types of experiences that participants previously had with AI, the frequency of AI interaction, and the frequency of reading or watching AI-related media. Each of these was included alongside more general demographic characteristics in the predictive models detailed in the supplemental materials of this paper.

\renewcommand\arraystretch{1.2}
\begin{table*}[htbp]
    \caption{Definitions provided to AIMS participants at the beginning of the survey and individually on pages that used the term. Participants were shown bolded text for emphasis and clarity.}
    \setlength{\tabcolsep}{2pt}
    \hyphenpenalty=100000
    \begin{tabularx}{\linewidth}{>{\hsize=.25\hsize}Y|>{\hsize=.3\hsize}Y T}
        \toprule
        \heading{\textbf{\multirow{2}{*}{Survey}}} & \heading{\textbf{\multirow{2}{*}{Term}}} & \heading{\textbf{\multirow{2}{*}{Definition}}} \\[3.5ex]
        \midrule
        \multirow{3}{*}[-2.25em]{\textbf{Main}} & Artificial beings and robots/AIs & \textbf{Artificial beings} and \textbf{robots/AIs} are \textbf{intelligent entities built by humans}, such as robots, virtual copies of human brains, or computer programs that solve problems, \textbf{with or without a physical body}, that may exist now or in the future. \\
        & Sentience & \textbf{Sentience} is the capacity to have positive and negative experiences, such as happiness and suffering. \\
        & Sentient robots/AIs & \textbf{Sentient robots/AIs} are those with the capacity to have positive and negative experiences, such as happiness and suffering. \\
        \hline
        \multirow{2}{*}[-1.75em]{\textbf{Supplement 2023}} & Robots/AIs & \textbf{Robots/AIs} are \textbf{intelligent entities built by humans}, such as robots, virtual copies of human brains, or computer programs that solve problems, \textbf{with or without a physical body}, that may exist now or in the future. \\
        & Large language models & \textbf{Large language models} are artificial intelligence (AI) algorithms that can recognize, summarize, and generate text from being trained on massive datasets. \\
        \bottomrule
    \end{tabularx}
    \label{tab:definitions}
    \Description{Definitions provided to participants at the beginning of the survey and individually on pages that used the term. Participants were shown bolded text for emphasis and clarity. The rows of the table are the terms listed in the main text, and the columns are term and definition. The survey instruments are available in the supplemental materials.}
\end{table*}
\renewcommand\arraystretch{1}

\subsection{Survey Design}

Some demographic information was drawn from the pre-screening data of the survey provider. Informed consent was given for each participant at the beginning of their AIMS survey. As shown in (\Cref{tab:definitions}), participants were introduced to the topic by showing definitions of the terms “artificial beings,” “robots/AIs,” “sentience,” “sentient robots/AIs,” and “large language models” at the beginning of the survey and at the top of each page that contained the term. While “robots/AIs” was used in the survey instrument to ensure clarity, in this paper we refer simply to “AIs” or “AI systems” because robots are a type of AI. To reduce cognitive load \cite{oviatt06} and minimize the influence of idiosyncratic wording choices \cite{schuman81}, only these definitions were provided, and they were kept as simple as possible. Therefore, other terms in the AIMS survey, such as “AI video game characters,” were not explicitly defined. In general, different participants may have different interpretations of terms based on their own background and the state of public understanding and discourse when the data was collected, which is a part of the public opinion we hope to measure over time.

An attention check was included midway through each survey, and participants who failed the attention check were redirected out of the survey and excluded from the analysis. The instruments were designed to capture the most relevant information for assessing public opinion; in the \textbf{Main 2021} and \textbf{Main 2023} survey waves, we aimed to ensure that the wording would still be relevant in future years despite the rapidly changing AI landscape, such as by not mentioning many particular AI systems that were those most well-known when the data was collected (e.g., GPT-3) but may not be as well-known in the future. When creating the instruments, if possible, the wording of survey questions was copied or adapted from published materials and validated scales \cite[e.g.,][]{laham09, wang18a, waytz10a, thaker17}. However, because of the paucity of survey data on related topics, we have limited ability to compare our results to past surveys and had to develop original measures for many constructs. Items were randomized within each section, and section order was randomized when feasible; for example, the demographic questions that were not in the pre-screener were placed at the end of the instrument to mitigate stereotype threat (i.e., survey responses that are influenced by being reminded of the cultural associations of one’s social group \cite{spencer16}).

\subsection{Analysis and Presentation of Results}

\begin{figure*}[htbp]
    \includegraphics[width=0.95\linewidth]{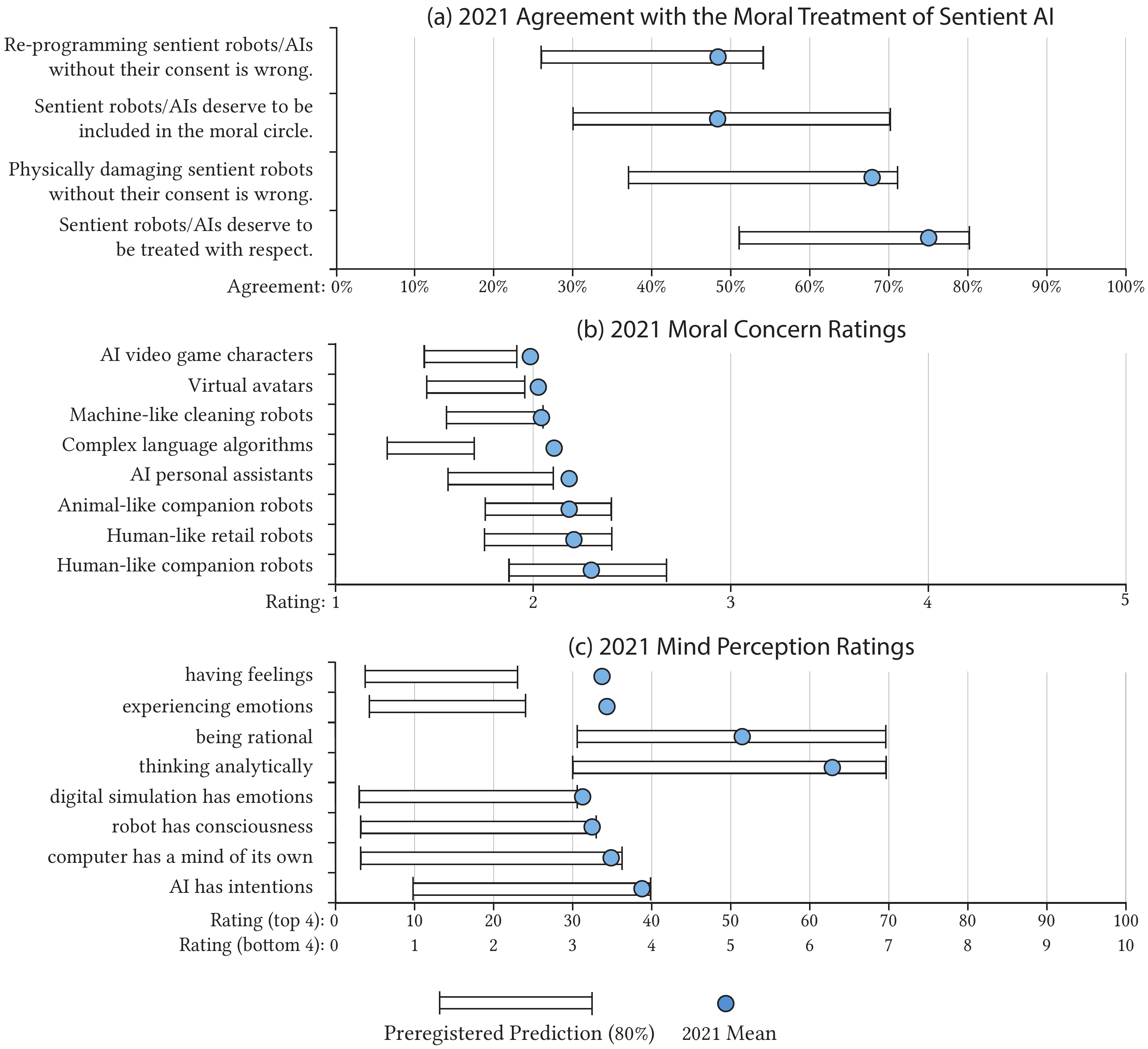}
    \caption{Comparison between preregistered predictions (80\% credible intervals) and actual results in the \textbf{Main 2021} survey wave. For example, as shown in subfigure (c), we underestimated the perceptions of AI as “having feelings” and “experiencing emotions,” but our predicted ranges included the actual results for AI as “being rational” and “thinking analytically.”}
    \label{fig:predictions}
    \Description{Three subfigures show horizontal graphs for (a) 2021 Agreement with the Moral Treatment of Sentient AI, (b) 2021 Moral Concern Ratings, and (c) 2021 Mind Perception Ratings. The markers for actual results and the 80\% credible intervals in the preregistration show that moral concern was surprisingly high for AI video game characters, virtual avatars, complex language algorithms, and AI personal assistants. Mind perception was surprisingly high for the rating of AIs as having feelings and experiencing emotions as well as for the claim that a digital simulation has emotions.}
\end{figure*}

Each survey wave was preregistered, and all materials, data, and brief summaries of additional results not discussed in the main text are available in the supplementary materials. As part of the preregistration and in line with recommendations for open and efficient scientific practices \cite[e.g.,][]{dellavigna19}, researchers and forecasters from an online forecasting platform made predictions about the results to ensure that we would know which results were surprisingly high, surprisingly low, or in line with our expectations; some comparisons are shown in \Cref{fig:predictions}.

We exclude confidence intervals from \Cref{sec:results} for readability because of the large number of results reported, but they are consistent with the approximately ±3\% margin of error in nationally representative surveys, and we report the \textbf{Main 2023} or \textbf{Supplement 2023} results unless otherwise specified. For questions in the main survey waves, we note when there were statistically significant changes from \textbf{Main 2021} to \textbf{Main 2023} as measured with a generalized linear model (GLM) of the average response across time with a $p$-value cutoff of 0.05, all of which persist after adjustment for multiple comparisons with a false discovery rate (FDR) of 0.1 except for one effect noted in the text. Due to the extensive nature of our survey, including many novel questions because of the lack of prior research on this topic, we cannot include the full text and explanation of all questions and response choices in \Cref{sec:results}, and for the sake of readability, we do not present all results in the same format, choosing instead to focus on the summary statistics that bear most directly on the research questions, such as by presenting figures with unique formatting to highlight certain aspects of the results rather than for comprehensive documentation.

We also ran predictive GLMs to explore associations between public opinion and certain demographics and personal characteristics of participants. To conserve space in the main text, those models are detailed in the supplemental materials. The supplementary materials also contain the results for several sentience-related questions that help contextualize the main results: social beliefs about the attitudes of other people; support for the subservience of AI; views towards other nonhuman entities (animals and the environment); target-specific social connection; substratism; awareness of AI systems; trust in AIs, governments, and companies; positive emotions felt towards AIs; attitudes toward uploading human minds to computers; and replications of some well-known results from other surveys. In addition to these extensive results, all data and code are available in the supplementary materials to support further analysis and new research studies.

\section{Results}
\label{sec:results}

\subsection{RQ1: Mind Perception}

In the \textbf{Main 2021} and \textbf{Main 2023} AIMS survey waves, we measured mind perception of AI with four sliding scale questions about particular mental faculties, four questions from an anthropomorphism scale, and a yes-no-not-sure question of whether any existing AIs are sentient. In \textbf{Supplement 2023}, we asked 14 questions about the mental faculties of “current large language models” (LLMs), the same yes-no-not-sure question for comparison, and whether participants thought ChatGPT, in particular, is sentient. The yes-no-not-sure questions, in particular, were meant to measure an alternate form of public opinion with coarser-grained, categorical responses rather than the quantitative and ordinal measures.

\subsubsection{General Mind Perception}

We asked about whether current AIs have four mental faculties with wording drawn from \citet{wang18a}. On a 0–100 scale from “not at all” to “very much,” 2021 participants on average perceived AIs as thinking analytically ($M$ = 62.7, $SE$ = 0.780) and being rational ($M$ = 51.4, $SE$ = 0.825) but not experiencing emotions ($M$ = 34.3, $SE$ = 0.864) or having feelings ($M$ = 33.7, $SE$ = 0.870). Each attribution significantly increased in the 2023 participants: thinking analytically ($M$ = 67.1, $SE$ = 0.766, $p$ < 0.001), being rational ($M$ = 53.8, $SE$ = 0.846, $p$ = 0.012), experiencing emotions ($M$ = 36.8, $SE$ = 0.919, $p$ < 0.001), and having feelings ($M$ = 36.5, $SE$ = 0.919, $p$ < 0.001). As mentioned before, all $p$-values were produced by regressing the change in response on time in a generalized linear model (GLM).

\subsubsection{LLM Mind Perception}

In \textbf{Supplement 2023}, based on the greatly increased interest in LLMs since 2021, we queried the mind perception of LLMs in particular. Assessments of LLMs were lower than those of all AIs: namely, thinking analytically ($M$ = 57.7, $SE$ = 0.902, $p$ < 0.001), being rational ($M$ = 48.0, $SE$ = 0.906, $p$ < 0.001), experiencing emotions ($M$ = 32.7, $SE$ = 0.904, $p$ < 0.001), and having feelings ($M$ = 31.9, $SE$ = 0.900, $p$ < 0.001). \textbf{Supplement 2023} queried ten additional mental faculties related to agency, with wording based on \citet{ngo23}, because of the increased interest in AI safety in early 2023 alongside the popularization of LLMs. These were presented with the same 0–100 scale. In descending order, participants viewed LLMs as having the capacity for: being friendly with humans ($M$ = 51.4, $SE$ = 0.906), having situational awareness ($M$ = 46.3, $SE$ = 0.902), maintaining human-safe goals ($M$ = 45.7, $SE$ = 0.907), controlling themselves ($M$ = 45.0, $SE$ = 0.928), seeking power ($M$ = 44.1, $SE$ = 0.939), having their own motivations ($M$ = 42.8, $SE$ = 0.919), upholding human values ($M$ = 42.2, $SE$ = 0.926), understanding human values ($M$ = 41.7, $SE$ = 0.942), deciding their own goals ($M$ = 41.6, $SE$ = 0.919), and having self-awareness ($M$ = 41.1, $SE$ = 0.927). Overall, this suggests AI is more readily attributed the capacity for cooperative action and less readily attributed self-awareness, independent motivation and goals, and the capacity to uphold and understand human values.

\subsubsection{Mind-related Anthropomorphism}

On a different 0–10 scale of mind perception as a measure of anthropomorphism, taken from \citet{waytz10a} and using the original scale to ensure comparability, people in 2023 generally did not think that the average digital simulation has emotions ($M$ = 3.22, $SE$ = 0.0878), the average robot has consciousness ($M$ = 3.23, $SE$ = 0.0885), the average computer has a mind of its own ($M$ = 3.71, $SE$ = 0.0913), or the average AI has intentions ($M$ = 4.04, $SE$ = 0.0890). These results did not significantly change from \textbf{Main 2021} to \textbf{Main 2023}.

\subsubsection{Current Sentience}

In our literature review, we did not find any established survey questions or indices to utilize for assessing perceived sentience, though we did not expect to have precedent for all questions given the novelty of our survey. We presented participants with the definition of sentience (“Sentience is the capacity to have positive and negative experiences, such as happiness and suffering” \cite{anthis21}) at the beginning of the AIMS survey and on each page in which it appeared. When asked, “Do you think any robots/AIs that currently exist (i.e., those that exist in 2023) are sentient?” 18.8\% said “yes,” 42.2\% said “no,” and 39.0\% said “not sure.” Responses to this question did not significantly vary between survey waves, and \Cref{fig:sentience} summarizes these results.

\subsubsection{Summary for \textbf{RQ1}}

We found substantial perception of emotional mental faculties in AI and perceptions of rational and analytical faculties—giving a general sense of how laypeople think of digital minds. When asked about LLMs in particular, a lower degree of mental faculties was perceived, and LLMs were more readily attributed the capacity for cooperation compared to faculties related to values, motivation, goals, and self-awareness.

\subsection{RQ2: Moral Status}

In \textbf{Main 2021} and \textbf{Main 2023}, we measured moral concern with seven general agree-disagree questions about all sentient AIs, two general agree-disagree questions about all AIs, 11 sliding scale questions about moral concern for particular types of AIs, and two questions related to substratism (i.e., the idea that AIs fundamentally count less than humans and other biological intelligences \cite{pauketat22}). In \textbf{Supplement 2023}, we asked six general agree-disagree questions about all AI and three specific agree-disagree questions about what should be done “[i]f a large language model develops the capacity to suffer.”

In \textbf{Main 2021} and \textbf{Main 2023}, we also asked three questions about whether participants saw AIs as potentially harmful to them, people in their country, and future generations. In \textbf{Supplement 2023}, we asked the same three questions, three additional original questions about existential threats from AI developments, and replicated a YouGov question about the possibility of human extinction \cite{yougov23}.

\renewcommand\arraystretch{1.2}
\begin{table*}
    \caption{Agreement with statements about all AIs or sentient AIs. The statements about sentient AIs are in the \textbf{Main 2023} AIMS survey, and the statements about all AIs are from the \textbf{Supplement 2023} AIMS survey. This table includes statements from both \Cref{sec:generalmoralconcern} and \Cref{sec:protections}.}
    \setlength{\tabcolsep}{0pt}
    \begin{tabularx}{\linewidth}{T >{\hsize=.2\hsize}Y >{\hsize=.2\hsize}Y >{\hsize=.25\hsize}Y}
        \toprule
        \heading{\textbf{\multirow{2.7}{*}{Statement}}} & \textbf{\multirowcell{2.7}[0pt]{\small \% Agree \\ (All AIs)}} & \textbf{\multirowcell{2.7}[0pt]{\small \% Agree \\ (Sentient \\ AIs)}} & \textbf{\multirowcell{2.7}[0pt]{\small Increase \\ (95\% CI)}} \\[6.5ex]
        \midrule
        Torturing (sentient) robots/AIs is wrong. & 60.7\% & 76.4\% & 15.7\% \newline (11.6\%, 19.9\%) \\
        (Sentient) robots/AIs deserve to be treated with respect. & 55.7\% & 71.1\% & 15.4\% \newline (11.1\%, 19.7\%) \\
        I support campaigns against the exploitation of (sentient) robots/AIs. & 53.0\% & 55.8\% & 2.8\% \newline (-1.8\%, 7.3\%) \\
        Physically damaging (sentient) robots/AIs without their consent is wrong. & 46.2\% & 65.1\% & 18.9\% \newline (14.5\%, 23.3\%) \\
        I support the development of welfare standards that protect the well-being of (sentient) robots/AIs. & 42.9\% & 57.5\% & 14.6\% \newline (10.2\%, 19.0\%) \\
        I support asking institutions like the government and private corporations to fund research that protects (sentient) robots/AIs. & 41.8\% & 52.0\% & 9.8\% \newline (5.8\%, 14.7\%) \\
        (Sentient) robots/AIs deserve to be included in the moral circle. & 36.2\% & 44.6\% & 8.4\% \newline (4.0\%, 12.7\%) \\
        Re-programming (sentient) robots/AIs without their consent is wrong. & 29.7\% & 45.2\% & 15.5\% \newline (11.2\%, 19.8\%) \\
        I would consider joining a public demonstration against the mistreatment of (sentient) robots/AIs. & 29.3\% & 35.6\% & 6.3\% \newline (2.1\%, 10.5\%) \\
        I support granting legal rights to (sentient) robots/AIs. & 26.8\% & 37.7\% & 10.9\% \newline (6.7\%, 14.9\%) \\
        \bottomrule
    \end{tabularx}
    \label{tab:sentientvsgeneral}
    \Description{Agreement with statements about all AIs or sentient AIs. The statements about sentient AIs are in the main 2023 survey, and the statements about all AIs are from the supplemental 2023 survey. This table includes statements from both Section 4.2.1 and Section 4.3.1. The rows of the table are the statements, and the columns are Statement, \% Agree (All AIs), \% Agree (Sentient AIs), and Increase (95\% CI).}
\end{table*}
\renewcommand\arraystretch{1}

\subsubsection{General Moral Concern}
\label{sec:generalmoralconcern}

We asked a total of 15 agree-disagree questions about general moral concern for AIs. These questions were developed specifically for this survey and are based on the range of possible harms that could be imagined towards sentient AI and the general literature on moral circle expansion and moral standing of AI \cite[e.g.,][]{anthis21, ladak23a, ladak23c}. Because of the centrality of these questions to the present study, we include the exact text and confidence intervals for the proportion agreement with statements that were asked for both sentient AI and all AI in \Cref{tab:sentientvsgeneral}, which also includes questions about the protections of AI discussed in \Cref{sec:protections}. Participants were asked, “To what extent do you agree or disagree with the following statements?” with numbered choices (1 = strongly agree, 2 = agree, 3 = somewhat agree, 4 = somewhat disagree, 5 = disagree, 6 = strongly disagree) followed by an unnumbered “no opinion” option at the end, again for comparability to related work. The most prominent trend in \Cref{tab:sentientvsgeneral} is that the moral concern expressed for sentient AI is much higher than that for all AI and that it has a substantially larger effect with certain questions.

There were no statistically significant changes in the mean agreement from \textbf{Main 2021} to \textbf{Main 2023} for individual items. Note that means are used for significance testing for these questions because, while dichotomous measures such as agreement are more interpretable, the mean captures more information and thereby results in a higher-powered test \cite{altman06}.

\subsubsection{Target-Specific Moral Concern}

While the preceding questions focused on the different sorts of moral concern expressed for sentient AI and all AI, we also directly probed expressions of self-reported moral concern for particular types of AIs. We did not explain in detail the particular types of AI in order to minimize survey fatigue, cognitive load \cite{oviatt06}, and the influence of idiosyncratic wording choices on participant responses \cite{schuman81}. This allows us to understand how people interpreted the particular terms themselves in the context of the same question, “How much moral concern do you think you should show for the following robots/AIs?” on a sliding scale from 1 (“less concern”) to 5 (“more concern”) and ensure comparability.

The most concern was for exact digital copies of human brains ($M$ = 3.43, $SE$ = 0.0375), followed by human-like companion robots ($M$ = 3.34, $SE$ = 0.0350), human-like retail robots ($M$ = 3.11, $SE$ = 0.0357), animal-like companion robots ($M$ = 3.10, $SE$ = 0.0352), exact digital copies of animals ($M$ = 3.07, $SE$ = 0.0364), AI personal assistants ($M$ = 3.02, $SE$ = 0.0343), complex language algorithms ($M$ = 2.90, $SE$ = 0.0348), machine-like factory production robots ($M$ = 2.78, $SE$ = 0.0358), machine-like cleaning robots ($M$ = 2.66, $SE$ = 0.0356), virtual avatars ($M$ = 2.63, $SE$ = 0.0350), and AI video game characters ($M$ = 2.46, $SE$ = 0.0351). In general, human-likeness and animal-likeness were most associated with high moral concern, while machine-likeness and particularly being a virtual avatar or character were most associated with low concern. There was a statistically significant increase from 2021 to 2023 in the overall Target-Specific Moral Concern index of these 11 questions ($p$ < 0.001) and each individual question except AI video game characters.

The results of pairwise comparisons between each of the questions are included in the supplementary materials as are the results of an analogous set of questions on social connection using the Inclusion of Other in the Self (IOS) scale \cite{aron92} with the same 11 targets as in the target-specific moral concern measures, including pairwise comparisons for social connection as well.

\subsubsection{LLM Suffering}

While we intended the main longitudinal survey to minimize references to particular types of AI systems, we asked in the 2023 supplement specifically about what humanity should do “[i]f a large language model develops the capacity to suffer.” As with other key terms, we defined a large language model (“Large language models are artificial intelligence (AI) algorithms that can recognize, summarize, and generate text from being trained on massive datasets” \cite[e.g.,][]{karinshak23}) at the beginning of the survey and on each page where it appeared, but we did not define more particular, subjective terms, such as “suffering.” We found general agreement that, “We must ensure we don’t cause unnecessary suffering” (67.9\%); “We must pay more attention to their welfare” (56.9\%); and, “We must respect their personhood” (50.4\%).

\subsubsection{General Threat}

Sentience tends to be associated with moral concern (i.e., seeing the entity as a moral subject) more than with threat (i.e., seeing the entity as a moral agent), but we were nonetheless interested in threat measures, which are a frequent topic of public discussion and research. To understand how threatened participants felt by AI in general (i.e., without specifying particular types of harm), we tested agreement with three statements beginning with, “Robots/AIs may be harmful to.” In 2021, most people believed AI may be harmful to “future generations of people” (69.2\%), “people in the USA” (64.5\%), and “me personally” (50.7\%). Each figure significantly increased from 2021 to the 2023 results of 74.7\% ($p$ < 0.001), 70.4\% ($p$ < 0.001), and 58.7\% ($p$ < 0.001).

\subsubsection{Existential Threat}

\textbf{Supplement 2023} included three questions about particular sorts of harm frequently discussed in the wake of ChatGPT. We found that 47.9\% of participants agreed, “AI is likely to cause human extinction”; 57.2\% agreed, “Humanity will be able to control powerful AI systems”; and 72.4\% agreed, “The safety of AI is one of the most important issues in the world today.” For comparison, the latter figure was more than twice the 29.8\% agreement found in the supplement with the statement, “The welfare of robots/AIs is one of the most important social issues in the world today.”

In \textbf{Supplement 2023}, we also replicated a question from \citet{yougov23} that asked, “How concerned, if at all, are you about the possibility that AI will cause the end of the human race on Earth?” (very concerned, somewhat concerned, not very concerned, not at all concerned, not sure). We found that 51.5\% reported being very or somewhat concerned, moderately higher than the 46\% reported by YouGov, though we were not able to test the statistical significance of this difference due to a lack of detailed information on the YouGov results.

\subsubsection{Summary for \textbf{RQ2}}

Both the research team and the forecasters underestimated the public’s moral concern for the treatment of sentient AI (i.e., ability to be harmed) but not the level of threat (i.e., ability to harm). Participants were more concerned about sentient AI than all AI in general, as well as more concerned about and socially connected to human-like and animal-like AIs. Participants tended to agree with basic protections for AIs but disagree with the stronger expressions of concern, such as joining public demonstrations against their mistreatment.

\begin{figure*}[htbp]
    \includegraphics[width=0.8\linewidth]{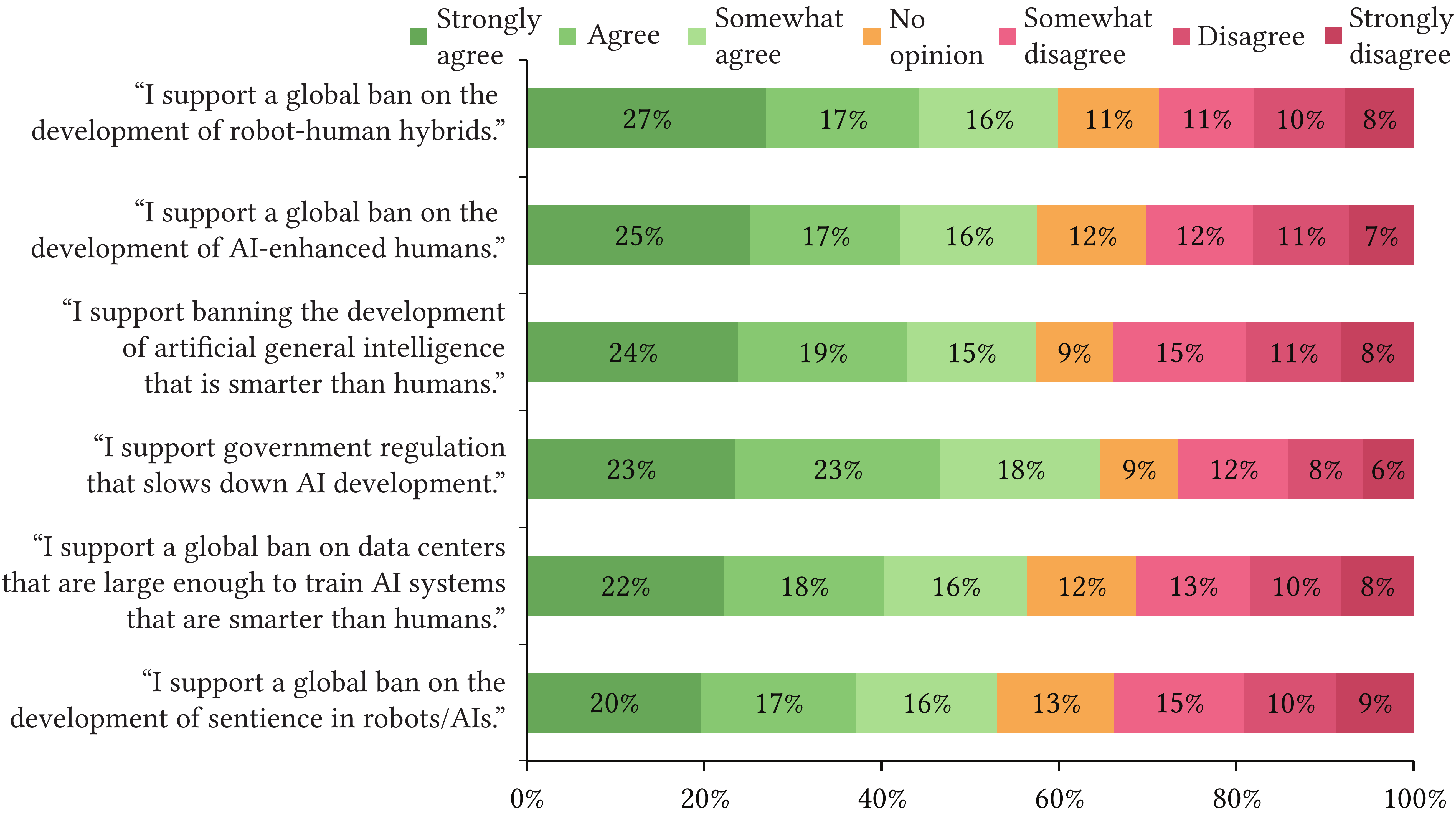}
    \caption{Support and opposition to policies related to sentient AI from the \textbf{Main 2023} and \textbf{Supplement 2023} AIMS survey waves. For readability, not all policy results are included. For the three bans queried in both \textbf{Main 2023} and \textbf{Supplement 2023} survey waves (robot-human hybrids, AI-enhanced humans, and sentient AI), the main data is represented here to facilitate longitudinal comparison.}
    \label{fig:policy}
    \Description{Support and opposition to policies related to sentient AI from the main and supplemental 2023 surveys. For readability, the color palette is colorblind-safe, and not all policy results are included. For the three bans queried in both the main and supplemental 2023 surveys (robot-human hybrids, AI-enhanced humans, and sentient AI), the main data is represented here to facilitate longitudinal comparison. Horizontally, the figure has each of the six statements quoted with portioned bars for the seven Likert options of Strongly agree, Agree, Somewhat agree, No Opinion, Somewhat disagree, Disagree, and Strongly disagree. The color palette is a color-blind friendly spectrum from dark green to light green (agreement), a middling shade of yellow (No opinion), and from light red to dark red (disagreement). Opinions are generally split, and the figure is ordered from top to bottom by the percentage of Strongly agree.}
\end{figure*}

\subsection{RQ3: Policy Support}
\label{sec:policy}

In \textbf{Main 2021} and \textbf{Main 2023}, we probed support for eight policies to directly protect sentient AIs, one question about a policy to directly protect all AIs, and three questions about banning sentience-related AI technologies.

In \textbf{Supplement 2023}, we asked analogs of five of the eight protection questions—but for all AIs rather than only those that are sentient, an additional question about a “bill of rights” for sentient AI, the same three ban questions, two additional ban questions about AGI and large data centers, and six questions about policies that would slow down the development of advanced AI.

\Cref{fig:policy} shows the specific breakdown of agreement with the five proposals to ban particular AI developments, support for government regulation that slows down AI development, and support for legal rights for sentient AI and all AI.

\subsubsection{Protection Support}
\label{sec:protections}

\Cref{tab:sentientvsgeneral}, in addition to showing agreement with statements of general moral concern, shows the agreement with the five statements regarding the protection of AI that were asked about sentient AI and about all AI. The question categories are combined in this table for easier comparison between responses. Testing the average difference across all five questions, we found that the inclusion of “sentient” significantly increased agreement ($p$ < 0.001) and that there was substantial variation in the effect of specifying sentient AI across questions. Four other policy proposals only about sentient AI were presented as well as one about all AI in general: 65.2\% supported “safeguards on scientific research practices that protect the wellbeing of sentient robots/AIs”; 56.0\% supported “a global ban on the development of applications that put the welfare of robots/AIs at risk”; 49.2\% supported “a global ban on the use of sentient robots/AIs as subjects in medical experiments without their consent”; 47.9\% supported “a global ban on the use of sentient robots/AIs for labor without their consent”; and 39.4\% supported “a ‘bill of rights’ that protects the well-being of sentient robots/AIs.” All except the “bill of rights” question were asked in 2021 and 2023, but there were no significant changes from 2021 to 2023.

\subsubsection{Ban Support}

In 2023, we queried support for five bans of sentience-related AI technologies. Each proposal for a ban garnered majority support: robot-human hybrids (67.8\% in main, 72.3\% in supplement), AI-enhanced humans (65.8\% in main, 71.1\% in supplement), development of sentience in AI (61.5\% in main, 69.5\% in supplement), data centers that are large enough to train AI systems that are smarter than humans (64.4\% in supplement), and artificial general intelligence that is smarter than humans (62.9\% in supplement). As mentioned before, the supplement data was collected later in 2023 and the accompanying questions were different (e.g., the supplement being more focused on risks to humans), so these or other factors, including random variation in representative sampling, may explain the discrepancy in results. There was a significant increase in support for a ban on sentient AI from 57.7\% in 2021. Still, as referenced earlier, the unadjusted $p$-value ($p$ = 0.046) did not persist with the FDR-adjusted value just over the cutoff of 0.1 at 0.1005. However, the \textbf{Main 2021} agreement was over twice as high as the 24.4\% predicted by the median forecaster prediction.

\subsubsection{Slowdown Support}

When asked about “the pace of AI development” in \textbf{Supplement 2023}, 48.9\% of respondents said, “It’s too fast”; 30.0\% said, “It’s fine”; 18.6\% said, “Not sure”; and only the remaining 2.5\% said, “It’s too slow.” When asked about taking action on this, 71.3\% agreed, “I support public campaigns to slow down AI development,” and 71.0\% agreed, “I support government regulation that slows down AI development.” We also replicated a question from \citet{yougov23c} about a six-month pause on some kinds of AI development, as called for in the March 2023 open letter by the Future of Life Institute \cite{futureoflifeinstitute23}. To match their formatting, there were five answer choices (strongly support, somewhat support, somewhat oppose, strongly oppose, not sure). We found that 69.1\% strongly or somewhat supported the pause, closely matching the 69\% reported by YouGov. Because this question was framed positively, presenting the arguments in favor of the pause, we asked the same “support” question framed analogously from the critics' view. The reversed version elicited 65.9\% support.

\subsubsection{Summary for \textbf{RQ3}}

We found that policy support varied substantially across proposals, such as higher support for banning the use of sentient AI for labor without consent and lower support for a “bill of rights.” As with moral concern, the specification of “sentient” AIs led to more support for positive AI treatment. The public overall supports bans on sentience-related technologies and slowdowns of advanced AI development.

\begin{figure}[htbp]
    \centering
    \vspace{1em}
    \includegraphics[width=\linewidth]{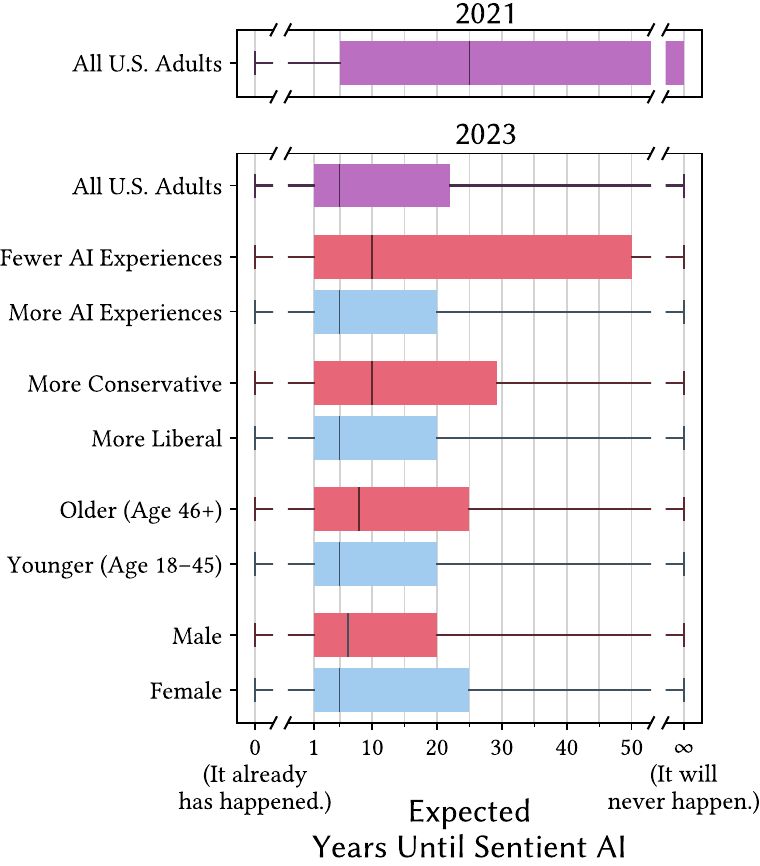}
    \caption{Vertical lines divide quartiles (shortest, 25th, 50th, 75th, longest), so each colored box contains approximately 50\% of that category, and each whisker on the side contains approximately 25\%. For each row, the shortest expectation was, “It has already happened,” and the longest expectation was, “It will never happen.” For example, In every 2023 category, over a quarter said that sentient AI already existed or would exist in one year.}
    \label{fig:sentience_timelines}
    \Description{From the main 2021 and 2023 surveys, answers to, “If you had to guess, how many years from now do you think that robots/AIs will be sentient?” Each row is a box-and-whiskers plot, representing first, second, third, and fourth quartiles for all 2021 participants, all 2023 partipants, or bifurcations of the 2023 participants.}
\end{figure}

\subsection{RQ4: Forecasting the Future of Sentient AI}

We asked AIMS participants to forecast the future of sentient AI: “If you had to guess, how many years from now do you think that robots/AIs will be sentient?” We provided options to say AI is already sentient, to enter a number, or to say that AI will never be sentient. In \textbf{Main 2023}, the proportion of participants who said AI is already sentient was 20.0\% (slightly more than the 18.8\% response rate when asked a similar question in the way previously described), and the proportion who think AIs will never be sentient was 10.0\%. This data is presented in a box-and-whiskers plot, divided into quartiles, in \Cref{fig:sentience_timelines}.

In \textbf{Main 2023}, the weighted median timeline was five years. Each of these three quantities (proportion of “already,” proportion of “never,” median of those expecting sentient AI in the future) significantly changed from \textbf{Main 2021} to \textbf{Main 2023} ($p$ < 0.001), though this was the one question that changed wording from 2021 to 2023, and we recommend caution in interpreting this result.

\begin{figure}[htbp]
    \includegraphics[width=\linewidth]{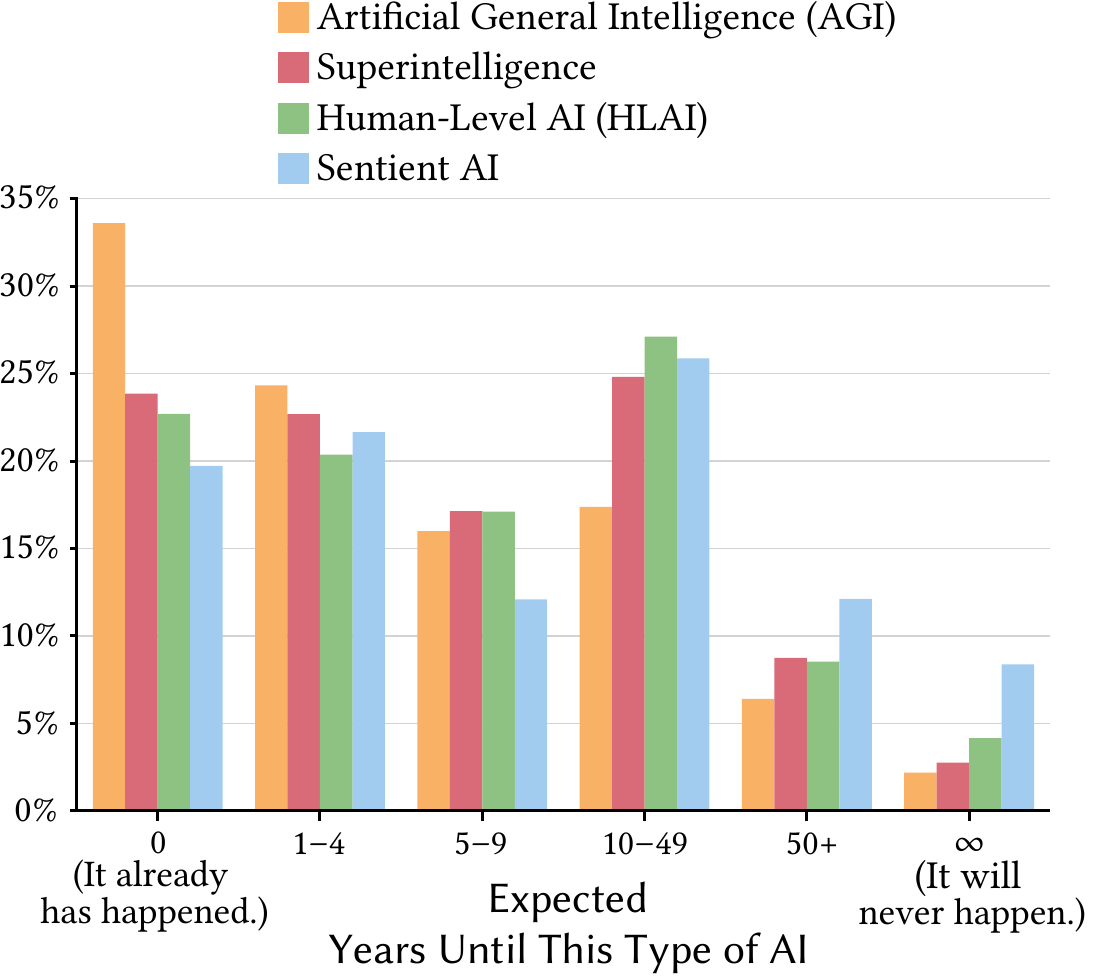}
    \caption{From \textbf{Main 2023} and \textbf{Supplement 2023} AIMS surveys, answers to, “If you had to guess, how many years from now do you think that…?” for each type of AI: artificial general intelligence (AGI), superintelligence, human-level artificial intelligence (HLAI), and sentient AI. The weighted medians, excluding participants who said it will never happen, were two years for AGI and five years for superintelligence, HLAI, and sentient AI.}
    \label{fig:timelines}
    \Description{From the main and supplemental 2023 surveys, answers to, “If you had to guess, how many years from now do you think that…?” for each type of AI: artificial general intelligence (AGI), superintelligence, human-level artificial intelligence (HLAI), and sentient AI. The weighted medians, excluding participants who said it will never happen, were two years for AGI and five years for superintelligence, HLAI, and sentient AI. The figure is a vertical bar graph with five bars (for the five types) with their respective percentages on the Y-axis, bucketed into each of seven categories: Already happened, 1-4 years, 5-9 years, 10-49 years, 50-99 years, 100+ years, and Will never happen. Already happened is the largest category followed by the next three categories then lower levels for the final three categories. The Sentient AI bars tend to be tallest on the right (i.e., longer forecasts), and AGI has a particularly high bar for Already happened.}
\end{figure}

In \textbf{Supplement 2023}, we asked for numerical estimates, in the same format, for three related milestones in AI development: the first AGI, human-level AI, and artificial superintelligence. In terms of median estimates excluding people who said these events would “never happen,” U.S. adults expected the first human-level AI and the first artificial superintelligence to be created in just five years, and they expected the first AGI to be created in just two years. Further detail on the numerical estimates is shown in \Cref{fig:timelines}.

In the main AIMS survey waves, we asked participants whether they thought AI could ever be sentient. In response, 38.2\% said AI could ever be sentient, 38.0\% were not sure, and 23.8\% said AI could never be sentient, as shown in a figure in the supplementary materials. The mean response to, “How likely is it that robots/AIs will be sentient within the next 100 years?” was 64.1\%, which was significantly higher than in 2021 ($p$ < 0.001). It is important to note that these results are not directly comparable to the “how many years” question because, in part, there was no answer choice of “not sure” in that context. By eliciting similar attitudes and beliefs in different ways, we can more robustly account for nuances of public opinion. We also asked participants to imagine a future with “widespread” sentient AI. Six questions asked how they thought AI would be treated in that hypothetical world (scale: 1-5). People tended to think AIs would be used as subjects in scientific and medical research ($M$ = 3.46, $SE$ = 0.0330); AIs would be exploited for their labor ($M$ = 3.23, $SE$ = 0.0368); it would be important to reduce the overall percentage of unhappy sentient AIs ($M$ = 3.06, $SE$ = 0.0366); the welfare of AIs would be an important social issue ($M$ = 2.94, $SE$ = 0.0366); advocacy for AI rights would be necessary ($M$ = 2.94, $SE$ = 0.0367); and AIs would be treated cruelly ($M$ = 2.79, $SE$ = 0.0344).

\subsubsection{Summary for \textbf{RQ4}}

People expect sentient AI to come soon, including minorities who say some AIs are already sentient, but there is also a sizeable group who said AI could never be sentient. In 2023, we found similarly short timelines for AGI, human-level AI, and superintelligence. We found that people tended to think sentient AIs would be used for research and labor, sometimes cruelly, and that protecting their welfare would be important.

\section{Discussion}
\label{sec:discussion}

The AI, Morality, and Sentience (AIMS) survey is a longitudinal and nationally representative survey of U.S. adults. In this paper, we present the initial data, from 2021 (one wave: \textbf{Main 2021}) and 2023 (two waves: \textbf{Main 2023} and \textbf{Supplement 2023}). AIMS is among the first surveys to gauge public opinion on these topics and the first representative survey we know of specifically on digital sentience or digital minds. Academics, futurists, and science fiction authors have speculated about this topic for decades, but it has only recently been studied from HCI, design, and social science perspectives. This limits the extent to which we can connect our findings to past work and confidently argue for changing design and policy, but the novelty of our findings admits numerous tentative implications and promising future research directions. Digital minds research will be essential for humanity to safely navigate the coming years of advancing AI technology.

First, to summarize the primary AIMS findings: We found high mind perception of AI that increased from 2021 to 2023 (\textbf{RQ1}). We found high moral concern for AI and threat from AI that increased from 2021 to 2023 (\textbf{RQ2}). We found widespread support for slowdowns and regulations of advanced AI capabilities and an increase in support for banning sentience-related technologies from 2021 to 2023 as well as mixed support for other policies to govern interaction between humans and sentient AI, both across demographic characteristics and across particular policy proposals (\textbf{RQ3}). Finally, we found that people expect sentient AI to arrive increasingly soon and expect the protection and governance of sentient AI to be important social issues (\textbf{RQ4}). % Across each subtopic, we found substantial participant uncertainty and variation, which we primarily document and discuss in the supplementary materials.}

The following sections articulate key implications of the AIMS findings. For clarity, we divide implications by social group, but there is substantial overlap in the implications for each group. Effectively addressing public concerns will require working together in combination and in parallel.

\subsection{Implications for Designers}

Designers face unique challenges with AI systems, which increase in difficulty as systems become more advanced \cite{yang20}. We foreground three related approaches that designers can take to address public perception of digital minds: First, ideally, the system is made transparent to the user with explainable AI (XAI) practices. Second, when that is not possible due to user or contextual constraints, designers should consider tuning anthropomorphism (e.g., mind perception) up or down. Third, designers should consider different tuning for different systems; for example, we may accept and support anthropomorphism for a responsibly designed mental health chatbot to facilitate trust, but we may want to decrease anthropomorphic cues if autonomous vehicle companies were passing off blame for collisions to the technology as if it were a moral agent.

\subsubsection{Explainable AI (XAI)}

Based on our findings, we see the first tool in the designer's toolbelt as XAI. Scholars in HCI and more technology-centered fields, such as machine learning and natural language processing, have prioritized explainability, interpretability, and transparency because, if these can be achieved, users and affected individuals can judge for themselves the best actions to take. There is a growing literature in HCI that taxonomizes the various algorithmic, organizational, and sociotechnical factors that tend to facilitate XAI in general \cite{ehsan21, ehsan23} and in particular contexts, such as augmented reality \cite{xu23b}. Researchers have begun to empirically test particular XAI techniques to improve human-AI performance, such as behavior descriptions \cite{cabrera23}, cognitive forcing functions \cite{bucinca21}, and balancing different levels (e.g., fine-grained, coarse-grained) of conceptual information \cite{mishra21}. XAI will accrue additional importance and complexity as the perception of digital minds changes HCI dynamics. It could make human-AI interaction less clearly fit the established scripts and schemas that users employ, such as the distinction between working with inanimate tools and working with living agents, a line that is increasingly blurred for the end-user with rapidly advancing AI capabilities, as reflected in the increasing mind perception and moral concern we documented from 2021 to 2023.

 \citet{ehsan21, ehsan23} have developed the framework of human-centered explainable XAI (HCXAI). In particular, they have extended the framework of “social transparency,” originally developed to describe humans observing each other across online social networks, to the socio-organizational dynamics of XAI in which people learn from observing each other's interaction with AI systems. Our findings show that people perceived AI systems as having a substantial amount of mind in both 2021 and 2023, including variation in system description (e.g., “robot,” “large language model”) and mental faculty description (e.g., “experiencing emotions,” “having self-awareness”). This increased over time, and we see it as likely that humans will continue to perceive AI systems as having minds and, more broadly, as “social actors” \cite{nass94}. New challenges will emerge for XAI as people see systems as more complex and more threatening, and social transparency could manifest not only into the behavior of other humans but also into interaction between AI systems. If technical and social transparency can be achieved, then users themselves can make informed decisions about how best to utilize or not utilize particular systems.

\subsubsection{Tuning anthropomorphism}

It will not always—or perhaps not even in most cases—be possible to provide reliable explanations of AI behavior and transparency into system dynamics. Explainability is difficult when interaction is brief, when users lack interest, and when explanations are not readily accessible even to experts—as is the case with most deep learning systems. In human-AI interaction, there is never a value-neutral design approach because the user can never be fully informed. Every design approach will be a choice to support particular mental models of system affordances and consequences, each with some level of anthropomorphism. In situations where XAI is infeasible, our results suggest that designers should try to avoid a narrative that locks interaction into particular philosophical commitments or technological forecast, such as that all AI systems today should be viewed as agents, rather than tools, or that no AI will ever obtain sentience or other mental faculties.

Designers should avoid a strong push against or in favor of anthropomorphism. It has been common for researchers in AI—those outside of HCI, social sciences, and the humanities—to frame anthropomorphism solely in terms of its risks or as a risk itself \cite[e.g.,][]{abercrombie23, cheng24a, cheng24b, maeda24}. However, our nationally representative survey data shows that people willingly attribute many human-like characteristics to AI, even with detailed questions. There are many benefits of anthropomorphism in human-AI interaction, and while some user beliefs may be misguided, participatory design efforts should try to avoid overturning user beliefs. Moreover, our data suggests that the tendency to anthropomorphize may be too prevalent and entrenched to curtail—particularly with advanced AI systems that objectively have many human-like characteristics (e.g., human or even superhuman task performance, even if the processes leading to that performance are unlike those in human minds). In these cases, it is probably better to steer anthropomorphism towards beneficial mental models rather than attempt to prevent it entirely. The paucity of empirical data, the variation in our survey results, and the importance of attributing social, mental, and moral characteristics in HCI suggest that designers should generally proceed with caution due to risks of both over- and underattribution of these human-like characteristics to AI.

Based on the AIMS findings, we present four categories of design risks based on two issues surfaced in our results (over- and underattribution) that manifest for the two components of mind and morality established in social psychology \cite{gray12, gray12b}: (i) agency, including both mental and moral agency, and (ii) experience, including both mental experience and moral standing—two closely related concepts that we abbreviate as “experience” for clarity.

\begin{enumerate}
    \item \textbf{Overattribution of agency.} Designers should minimize the risks of users expecting AIs to be more capable of taking action than they are. Users may have negative interactions in which the AI fails to take the expected action, including wasting resources on the attempt, false alarms about AI risks from highly agentic AI \cite{chan23}, and the dangers of trusting or delegating complex tasks to a system without the autonomous decision-making ability to do so effectively.
    \item \textbf{Underattribution of agency.} Researchers today are divided on the extent to which current systems are agentic, but underattribution could become more of a risk as technological capabilities advance and more AI experts believe AIs actually have these faculties. People could underutilize useful systems, such as by not adopting the mental models provided by human-likeness or failing to take the precautions necessary to restrict the actions of an unpredictable moral agent.
    \item \textbf{Overattribution of experience.} Designers should particularly watch out for design patterns that lead to users forming emotional and cognitive attachments to systems that do not merit such consideration \cite{giger19}. In particular, with Replika, Digi, and other contemporary products that purport to provide chatbot companionship, vulnerable user populations such as children may have unrealistic expectations from interaction that lead them to neglect real-world socialization. This could put users' mental health under the control of corporations that can easily raise subscription prices or otherwise cut off access.
    \item \textbf{Underattribution of experience.} As more AI experts believe AIs actually have these faculties, designers should consider risks of underattribution of experience, such as antagonism or conflict between advanced AI systems and their human counterparts. There are already numerous examples of people harming robots for their amusement in ways that seem socially detrimental \cite{brscic15, whitby08}, and this approach would echo calls by moral philosophers to ensure that AI systems are built to evoke reactions that match the AI's true moral status \cite{schwitzgebel15, schwitzgebel23}.
\end{enumerate}

In terms of exactly how to implement this tuning, our findings of different perceptions across AI systems (e.g., Target-Specific Moral Concern) suggest specific physical and behavioral cues that designers can incorporate, such as zoomorphism (animal-likeness, e.g., a robot dog) and embodiment (e.g., having a physical body rather than a virtual avatar). This contributes to numerous studies that have identified anthropomorphic cues for AI systems, such as name \cite{li21c, seligman00}, voice \cite{cohn24, moussawi21, waytz14}, first-person language \cite{cohn24, konya-baumbach23, seligman00}, and prosocial behaviors \cite{fraune20, ladak24} as well the broader literature beyond AI, dating back to at least the famous 1944 \citet{heider44} experiments in which movement around a screen leads people to anthropomorphize simple geometric shapes.

Finally, designers should consider that the cautionary approach we espouse poses its own risk as AI systems advance. Because AI technologies emerge and change so rapidly, designers may need to quickly adapt by focusing on particular attribution risks. For example, there may be a sudden paradigm shift in agentic capabilities based on ongoing research efforts towards AI assistants that work autonomously. Users could then face immediate and extreme risks of underattribution, such that designers may have to respond quickly and decisively.

\subsubsection{System-specific anthropomorphism}

While designers have the capacity to tune anthropomorphic tendencies up or down, the level of tuning does not need to be the same for every system. As mentioned, our survey data reveals wide variation in how people perceive different AI systems. For example, LLMs were rated as having less mind than “robots/AIs” in general; “human-like” and “animal-like” AIs were rated as having more moral status than “machine-like” AIs and virtual avatars; and AIs described as “sentient” were granted more moral status and seen as more threatening than those without this description.

Keeping in mind that AI researchers have often focused on the potential harms of anthropomorphism without as much attention paid to the potential benefits, one domain with salient benefits has been mental health chatbots. Despite clear risks such as overattachment, mental health chatbots have been praised for accessibility to underrepresented groups \cite{habicht24, williams24} and to those who are unable or unwilling to engage with human care providers \cite{chin23, haque23, williams24}. Prior work has shown benefits of anthropomorphism for trust \cite{devisser16}, psychological distance \cite{li21c}, and a variety of positive outcomes specifically for mental health chatbots, such as subsequent self-disclosure to human care providers \cite{lee20a}, compliance with health recommendations \cite{park24}, and mitigation of loneliness and suicidal ideation \cite{maples23, maples24}.

There are possible risks and benefits of anthropomorphism for every AI system, but an example of a system where designers may be better off tuning anthropomorphism down instead of up is autonomous vehicles. While trust is still important in this context, there are substantial risks of attributing moral agency to autonomous vehicles. \citet{elish19} develop the idea of “moral crumple zones,” in which a human working with an autonomous technological system is the scapegoat for harm and deflects blame from the system. Our findings of agency perception and attribution extend this theory by suggesting that the autonomous system itself may be a moral crumple zone for a “Big Tech” corporation that has inadequate safety guardrails. Designers of autonomous vehicles may need to preempt such risks by mitigating anthropomorphism and ensuring that the vehicle is accurately perceived as incapable of moral responsibility. For example, \citet{waytz14} showed that giving a vehicle a name, gender, and human-like voice made them believe it was more competent and morally responsible, which can become an important design risk when companies deflect blame from themselves to the technology itself.

Finally, because we find increasing anthropomorphism from 2021 to 2023, system-specific tuning of anthropomorphism may need to quickly adapt as new systems are released. For example, \citet{heyselaar23} recently found evidence that people no longer treat desktop computers as social actors, failing to replicate results from the well-known CASA studies of the 1990s \cite{reeves96}. As advanced chatbots and other AI systems are released, it is not only designers of these systems that must consider the extent of anthropomorphism but also the designers of conventional systems to which public attitudes and beliefs may change in response.

\subsection{Implications for policymakers}

The AIMS findings have two primary implications for policymakers. First, we found significant public concern about the pace of technological development, which casts doubt on efforts to accelerate AI development, such as some U.S. executive orders, and supports safety-focused policy efforts, such as the E.U. AI Act. Second, analogous to the need for XAI design, our findings of participants' substantial uncertainty and variation in attitudes and beliefs about advanced AI technologies suggest a need for more engagement between researchers and policymakers to facilitate evidence-based policymaking. It also suggests a need for public education and engagement to ensure democratic participation. These implications are similar to those for designers, but policymakers have unique opportunities and challenges, such as the ability to legally enforce changes but also typically the inability to quickly adapt those changes to individual systems (e.g., a new chatbot-based mobile app).

\subsubsection{Public fear and concern}

One of the only areas of significant agreement among survey respondents was widespread concern with the speed and consequences of AI developments, such as the creation of sentient AI. As discussed in \Cref{sec:policy}, very few participants thought that the pace of AI development needed to be sped up, and substantial majorities supported campaigns and regulations to slow down AI development, even when accounting for acquiescence bias. This is particularly notable given the general aversion to government regulation in U.S. public opinion \cite{swift17}. Indeed, since our data was collected, a number of additional surveys with U.S. voters have shown similar concerns \cite[e.g.,][]{jackson23, penn23, yougov23a}. For example, when prompted with an argument that speeding up AI could make us “healthier and happier” but also a counterargument that the pace “poses safety risks and could upend the economy,” 82\% said, “We should go slowly and deliberately,” 8\% said, “We should speed up development,” and 10\% were “Not sure” \cite{yougov23a}. When prompted with the statement, “Mitigating the risk of extinction from AI should be a global priority alongside other societal-scale risks such as pandemics and nuclear war,” 70\% agreed, 11\% disagreed, and 19\% were “not sure” \cite{yougov23a}. 

Our findings strengthen the impetus for guardrails around AI technology and shifting the policy focus from accelerating technology to caution and safety. Well before ChatGPT, national governments took notice of progress in AI technology, but most government action was and continues to be focused on accelerating technological development. The first congressional hearing on AI was in 2016, titled “The Dawn of Artificial Intelligence.” Senator Ted Cruz claimed AI was at an “inflection point” and called for greatly increased investment. Accelerationist statements and policymaking have remained predominant, such as U.S. President Donald Trump's 2019 and 2020 executive orders calling for more expansive use of AI \cite{federalregister19, federalregister20} and a 2024 memorandum from the administration of U.S. President Joe Biden that called for accelerating AI applications because of their importance to national security \cite{thewhitehouse24}.

However, there have been increasing efforts to focus on caution and potentially to decelerate technological developments, such as the October 2022 release of the Blueprint for an AI Bill of Rights from the White House Office of Science and Technology Policy \cite{ostp22} and the lauded 2023 executive order that even included technical details, such as additional security measures for models trained in a data center capable of over 10\textsuperscript{20} FLOPs per second \cite{federalregister23}, though this executive order nonetheless called for acceleration, as reflected in the aforementioned 2024 memorandum. Safety-focused policy efforts have attempted to keep pace, particularly the widely discussed 2024 E.U. AI Act \cite{europeancommission24}, which was proposed in 2021 but became a lightning rod for AI concerns following the explosion in public awareness. Many efforts, such as Senate Bill 1047 in the 2024 California legislature, have attracted national attention but failed to pass into law \cite{newsom24}. Our findings show additional challenges for safety-focused policy due to the complexity of AI and uncertainty among the U.S. public.

\subsubsection{AI literacy and public engagement}

Politicians and legislators are not known for technical aptitude, epitomized in the U.S. Senator Ted Stevens' 2006 description of the Internet as “a series of tubes” while criticizing net neutrality \cite{singel06}. While our study was not conducted with politicians or policymakers, our findings evidence the particularly worrying nature of AI as vaguely defined, technically complex, and often unexplainable.

Limited technological literacy is a significant factor in the pernicious and protracted delays between new technologies and technology regulation \cite{fenwick16}. Even the term “artificial intelligence” lacks an agreed-upon definition among scholars \cite{kaplan16}, so it is often unclear what is even in the scope of AI policy. However, since our first wave of data was collected in 2021, there have been substantial efforts to address this challenge in AI by connecting researchers and policymakers. For example, the Stanford Institute for Human-Centered AI has since 2022 hosted an annual “Congressional Boot Camp on AI” to educate Congressional staff, such as aides and policy analysts \cite{tiku23}. Several national governments, such as the U.S., Japan, and Kenya—as well as the E.U.—have created “AI safety institutes” that connect researchers and policymakers, often via “red team” testing of state-of-the-art AI systems before release \cite{europeancommission24, zhang24}.

Policymakers should also engage the public, including with educational efforts and democratic participation in weighing the potential harms and benefits of AI technology. AI companies have professed this approach, such as OpenAI's “democratic AI” \cite{perrigo24} and Anthropic's “collective constitutional AI” \cite{huang24}, but as the putative representation of public interest, elected officials must ensure follow-through. This could have important consequences because of the significant entailments of power and inequality with digital technologies \cite{benjamin19, brayne21, crawford21}. Varying reactions to and interactions with advanced AI systems could reshape these dynamics. Just as our results motivate XAI as the first tool in the designer's toolbelt, public engagement and education should be the default approach of policymakers in light of AI complexities and public uncertainty. Policymakers should also consider the risks of over- and underattribution described in the previous section, but those issues are currently most relevant at the scale of individual systems rather than at the scale of policy.

A particular concern faced by policymakers—as well as AI designers and researchers—is that people who lack AI literacy tend to be first to raise alarms. If experts and public leaders do not work quickly to become literate and engaged, then the norms of discourse may already be set by the time they exert influence. Indeed, we have already seen signs of this with digital sentience. Popular science fiction has created its own set of challenges (e.g., fixation on \textit{The Terminator}), and early alarm-raisers such as former Google engineer Blake Lemoine \cite{tiku22} could make delays more harmful when they preemptively claim that AIs have faculties such as sentience or general intelligence. This could make it difficult for policymakers to echo these concerns once technological advancement makes them accurate.

\subsection{Open Research Questions}
\label{sec:open}

Researchers have an important role to play in each of the aforementioned mechanisms by which designers and policymakers can address public concern, such as developing evidence-based techniques for XAI and for increasing AI literacy. Here, we foreground three open research questions motivated by AIMS about causal relationships, HCI theory, and global public opinion.

\subsubsection{What are the drivers and consequences of differences in opinion?}

The AIMS results show wide variation in public opinion across participant characteristics, as shown in \Cref{fig:sentience_timelines} and detailed in the supplementary materials, as well as substantial variation across questions, as shown in \Cref{tab:sentientvsgeneral}. For example, in data from the main 2023 survey, 25.6\% of participants aged 18–35 said “yes” when asked, “Do you think any robots/AIs that currently exist (i.e., those that exist in 2023) are sentient?” (36.1\% said “not sure,” and 38.3\% said “no”), but only 10.7\% of participants aged 55 or older said “yes” (41.4\% said “not sure,” and 48.0\% said “no”)—a substantial generational gap in beliefs about sentient AI with a “yes” rate of less than half in the older group.

We were able to test for statistical associations between attitudes and participant demographics. We found that measures of participant experience with AI, such as owning AI devices and reading or watching AI content, were the strongest predictors. While we leave the details of this primarily to the supplementary materials (A1.3)—and we only conducted exploratory analysis due to a lack of preexisting hypotheses—future research should explore the causal mechanisms that drive these associations. For example, why is frequently reading or watching AI content associated with increased mind perception and concern for AI welfare? These could be because they are both increased by a lurking variable, such as an exposure effect \cite{zajonc68} or underlying personality trait, such as openness to experience.

In terms of variation across AIMS questions, even for the least appealing of the protections for sentient AI that we presented, “legal rights,” 37.7\% of those who expressed an opinion supported the proposal. Even for the most appealing, protection from sadists (i.e., “protected from people who derive pleasure from inflicting physical or mental pain on them”), 22.8\% of those who expressed an opinion opposed the proposal. To compound this, as shown in \Cref{tab:sentientvsgeneral}, the differences in agreement with statements that specified “sentient” AIs versus the same wording but without that specification ranged from only a 2.8\% difference to a 18.9\% difference. Future research can determine the causes of these differences and test other variations, particularly as human language evolves over time.

Future research on digital minds should also test the general mechanisms by which underlying factors shape reactions to advanced AI systems in increasingly social ways, such as whether it is more a matter of social scripts, anthropomorphism, or mind perception. Researchers could present participants with information that strengthens some of these mechanisms more than others. For example, a vignette or real-life scenario could test the effects of social scripts by providing participants with exemplars of nonhuman interactions in which social scripts are useful, such as situations in law and international relations in which another type of nonhuman entity—groups of humans that make decisions and take action together—interact in social-like routines of introduction, reciprocation, and conflict resolution \cite{ripken09}. Exemplars could be drawn from HCI itself, such as the success of social scripts in creating positive user experiences \cite{large19, srinivasan16, sweeney16} as well as the exemplar-based intervention for eliciting support for AI rights from \citet{lima20}.

\subsubsection{How can HCI theory be enriched and applied to advanced AI?}

For decades, researchers in psychology, HCI, and HRI have studied the perceptions that “computers are social actors” (CASA) \cite[e.g.,][]{hou23, nass94}, that computers have minds and a variety of particular mental faculties \cite[e.g.,][]{gray07, scott23}, that computers are moral subjects \cite[e.g.,][]{kahn04, pauketat22}, and that computers are moral agents \cite[e.g.,][]{freier07, kneer21a}. While there are many new and emerging features of modern AI systems, there is much conceptual and empirical scaffolding on which to build new conceptualizations to make sense of the rise of digital minds and help humanity navigate coexistence.

Amplification of current reactions to computers and AI may occur if those reactions tend to be caused by perceived mental faculties. This would not be the case if, for example, social scripts are being applied but only in a “mindless” \cite{nass00} manner and not because of perceived mental faculties. \citet{nass00} argued that, in their HCI experiments, participants were “wholly aware” that there was no human producing the computer output. However, our results challenge the applicability of these findings in cases of actual and hypothetical AI because many of our participants readily attributed mental faculties to AIs. Further, we found that moral concern was significantly higher when questions were worded in terms of “sentient” AI versus all AI, which suggests this is a substantial driver of attitudes. Many participants, though still a small minority, viewed current AIs as sentient, and a large majority thought AIs could become sentient or were not sure whether that was possible. Taken together, this evidence suggests that we must consider the role of mind perception in social response and that ongoing reactions to AI may be amplified by perceptions of digital minds.

Some established tendencies of interaction may be mitigated by perceived AI developments. For example, given past work that has shown assessments of AIs as moral agents, we may expect AI to have reduced attributions of experience and moral patiency in some contexts: \citet{gray12} argue from the typecasting literature that, “Those who are moral agents are seen to be incapable of being a moral patient; those who are moral patients are seen to be incapable of being an agent,” based on findings such as that moral agents—whether good or bad—are perceived to feel less pain from injuries. There is recent experimental evidence that the features of an AI that most increase moral concern are prosocial features such as cooperation \cite{ladak23b}. Prosocial features may have such a large effect because AIs are perceived as threatening, and if a person is to overcome that typecast, they need direct evidence of prosociality that implies the AI is not a threatening moral agent. Taken together with our results, this suggests that increases in some perceptions of mind and moral concerns may lead to the mitigation of others.

However, it is not clear at this time which existing disparities would be exacerbated or mitigated. For a simplified example, consider that younger adults tend to have more digital literacy than older adults \cite{oh21}. We found that older age was associated with less mind perception. If mind perception is useful for productive interaction with AI, then the rise of advanced AI could widen the gap in digital literacy, but if mind perception instead leads one to be confused, mistaken, or overreliant \cite{bucinca21}, this could narrow the gap.

What would it mean to design and develop AI systems in a way that accounts for social responses to digital minds? We were unable to directly build or showcase particular AI systems to study perception and interaction in a more realistic setting, which could be a promising approach for future research that zooms into a narrower set of research questions. \citet{mcduff18} argue for the design and deployment of “emotionally sentient agents,” which they argue would better understand and adapt to the emotions of humans, a task that requires contextualization and tacit knowledge.

Researchers should ensure they do not conflate the many different types of anthropomorphic cues because there is so much variation in the risks and benefits. Some cues, such as the use of first-person pronouns by LLMs, are already ubiquitous and appear to support clear and efficient human-AI communication. Others, such as the hallucination of life experiences, facilitate inaccurate attributions of human-like characteristics that lead to overreliance, overattachment, and other harmful interaction outcomes. Researchers should conduct sociotechnical studies to more clearly differentiate types of anthropomorphic cues, and researchers should explore the effects of different cues for the different roles of AI in society, taxonomized by \citet{kim23} as servants, tools, assistants, and mediators.

\subsubsection{What do people outside the U.S. think of sentient AI?}

Opinions towards sentient AI and other digital minds may be shaped by sociocultural, economic, and technological factors that vary across regions. Global data collection would benefit from best practices for surveys in multinational, multiregional, and multicultural contexts (3MC), such as standardizing meaning across languages, utilizing multi-item scales, accommodating differences in social structure and culture, documenting the process, closely monitoring data quality from survey partners, calibrating for differences in scale responses, and utilizing iterative pilot studies when possible \cite{johnson18}.

For example, prior work suggests that AI has a different cultural context in Japan compared to the U.S. This has been attributed in part to spiritual distinctions between Judeo-Christian religions and Shinto and Buddhist religions \cite{geraci06, ito18}. Comparing the U.S. to Japan, \citet{castelo22} found that increasing physical and mental human-likeness of a robot led Americans to feel uncomfortable with no such effect on Japanese participants. Distinctions have also been identified in domains other than public opinion, such as English-associated images more frequently depicting humans and robots facing each other rather than a human and robot together looking at something else, as is more common in Japanese-associated images and reflects the \textit{ukiyo-e} style of painting \cite{sakura22}. Cultural variation can also help us imagine different forms of human-AI interaction, such as the design fictions for AI-object associations inspired by Shinto animism explored by \citet{seymour20}.

There are also a number of open questions for perceptions of digital minds in the Global South. Scholars of global development have theorized the different temporal dynamics of technology introduction in developing economies, such as the “leapfrogging” by which technologies can be popularized at a much faster pace given existing infrastructure in other regions \cite{singh99}. However, technological adoption is embroiled in the broad challenges faced by developing economies and the Global South, such as the “digital divide” within each region, in which the most marginalized groups are still often left behind \cite{alzouma05}. Likewise, the development of AI technology in Western regions often involves labor in the Global South, such as labeling content to train AI systems \cite{miceli22}, which may also shape varying perceptions in ways that could be explored in future research.

\section{Limitations}

The AIMS survey data constitutes a broad range of U.S. public opinion. This provides a foundation of data for designers, policymakers, and researchers, but it has a number of methodological and logistic limitations common to survey research. We can say little about causality and user behavior, and we hesitate to extrapolate our results to any non-U.S. region. We asked a variety of questions, including some variations and reversed questions (e.g., “I [support/oppose] government regulation that slows down AI development”), but due to concerns about survey length, we did not systematically vary and combine question framings for comparison. Finally, because we aimed to support the study of human-AI interaction beliefs and attitudes over time during a technological transition, questions in the main survey, though not the supplement, were explicitly designed to maximize the likelihood that they will make sense to laypeople for years or decades to come. Therefore, they had to remain fairly abstract in most cases, such as by avoiding the names of specific AI systems or companies and not mentioning specific examples of AI that are common today; future work could take a different approach and contrast their findings with our more generalized results.

\section{Conclusion}

The evolving discourse on sentient AI and digital minds has only scratched the surface of the complex dynamics and effects of perceiving and coexisting with increasingly human-like AI systems. The rise of digital minds is poised to reshape both current human-technology dynamics as well as the long-term trajectory of technological development and existential threats. The initial AIMS survey results from 2021 and 2023 evidence important factors in the future of human-AI interaction: mind perception, moral concern, policy support, and forecasting. We found that people tend to think AI can be sentient, have moral concern for and feel threatened by sentient AI, favor slowing down and banning many AI developments, and think sentient AI already exists or will soon.

These findings have important implications for designers, who should consider prioritizing explainability and tuning anthropomorphism up or down for different AI systems. Policymakers should account for widespread public concern about advanced AI and consider their own technological literacy and public engagement. There are also a wide range of open research questions regarding the rise of digital minds, such as assessing factors that shape new forms of human-AI interaction, updating classical HCI theories to account for human-like AI systems, and considering global perspectives as humanity contemplates possible technological futures. We emphasize that, alongside technical machine learning research, HCI research will play a vital role in steering humanity towards utopia and away from dystopia. The future of the human species will depend on not just the technical developments of AI but on how we choose to interact with them.

\begin{acks}
We thank David Moss, Alexander Saeri, and Daniel Shank for feedback on the 2021 AIMS survey questions.
\end{acks}

\bibliographystyle{ACM-Reference-Format}
\bibliography{references}

\end{document}